\documentclass[lettersize,journal]{IEEEtran}
\usepackage{amsmath,amsfonts}
\usepackage{soul}
\usepackage{algorithmic}
\usepackage{algorithm}
\usepackage{array}
\usepackage[caption=false,font=normalsize,labelfont=sf,textfont=sf]{subfig}
\usepackage{textcomp}
\usepackage{stfloats}
\usepackage{url}
\usepackage{verbatim}
\usepackage{graphicx}
\graphicspath{{img/}}
\usepackage{subfiles} 
\usepackage{cite}
\usepackage{colortbl}
\usepackage[table]{xcolor}
\usepackage{xcolor, soul}
\definecolor{LightBlue}{rgb}{0.68, 0.85, 0.98}
\sethlcolor{LightBlue}
\usepackage{threeparttable}
\usepackage{booktabs}
\usepackage{multirow}

\usepackage{titlesec}
\titleformat{\subsubsection}
{\normalfont\normalsize\bfseries}{\arabic{subsubsection})}{0.5em}{}
\begin{document}

\title{Vision-Language Models can Identify \\ Distracted Driver Behavior from Naturalistic Videos}

\author{Md Zahid Hasan, Jiajing Chen, Jiyang Wang, Mohammed Shaiqur Rahman, Ameya Joshi, \\ Senem Velipasalar, Chinmay Hegde, Anuj Sharma, Soumik Sarkar 
\thanks{M.Z. Hasan, M.S. Rahman, A. Sharma, and S. Sarkar are with
Iowa State University, Ames, IA-50010, e-mail: \{zahid, shaiqur, anujs, soumiks\}@iastate.edu

J. Chen, J. Wang and S. Velipasalar are with the Department of Electrical Engineering and Computer Science, Syracuse University, Syracuse, NY-13244, e-mail: \{jchen152, jwang127, svelipas\}@syr.edu

A. Joshi and C. Hegde are with the Department of Electrical and Computer Engineering, New York University, Brooklyn, NY-11201, e-mail: \{ameya.joshi, chinmay.h\}@nyu.edu
}

\thanks{This material is based upon work supported by the Federal Highway Administration Exploratory Advanced Research Program under Agreement No. 693JJ31950022. Any opinions, findings, and conclusions or recommendations expressed in this publication are those of the Author(s) and do not necessarily reflect the view of the Federal Highway Administration.}
\thanks{Manuscript received June 22, 2023; revised March 04, 2024.}}

\markboth{Journal of \LaTeX\ Class Files,~Vol.~, No.~, June~2023}%
{Shell \MakeLowercase{\textit{et al.}}: A Sample Article Using IEEEtran.cls for IEEE Journals}


\maketitle
\begin{abstract}
Recognizing the activities causing distraction in real-world driving scenarios is critical for ensuring the safety and reliability of both drivers and pedestrians on the roadways. Conventional computer vision techniques are typically data-intensive and require a large volume of annotated training data to detect and classify various distracted driving behaviors, thereby limiting their generalization ability, efficiency and scalability. We aim to develop a generalized framework that showcases robust performance with access to limited or no annotated training data. Recently, vision-language models have offered large-scale visual-textual pretraining that can be adapted to task-specific learning like distracted driving activity recognition. Vision-language pretraining models like CLIP have shown significant promise in learning natural language-guided visual representations. This paper proposes a CLIP-based driver activity recognition approach that identifies driver distraction from naturalistic driving images and videos. CLIP's vision embedding offers zero-shot transfer and task-based finetuning, which can classify distracted activities from naturalistic driving video. Our results show that this framework offers state-of-the-art performance on zero-shot transfer, finetuning and video-based models for predicting the driver's state on four public datasets. We propose frame-based and video-based frameworks developed on top of the CLIP's visual representation for distracted driving detection and classification tasks and report the results. Our code is available at \url{https://github.com/zahid-isu/DriveCLIP}
\end{abstract}

\section{Introduction}

\IEEEPARstart{D}{I}stracted driving accounts for 8\% of fatal crashes, 14\% of injury
crashes, and 13\% of all police-reported motor vehicle traffic crashes in 2021 on U.S. roads. It has emerged as a significant public safety concern and contributed to a substantial number of road accidents worldwide~\cite{NHTSA2023}. The World Health Organization reported that
1.19 million people died in traffic accidents worldwide in
2023~\cite{WHO2023}. Although cell phone usage is often synonymous with distracted driving, the problem is broader. The Center for Disease Control and Prevention (CDC) identifies three forms of distraction~\cite{cdc}: i) visual distraction, where the driver takes their eyes away from the road. This can be due to several reasons, such as looking at the GPS, Billboard, etc.; ii) manual distraction, where the driver takes their hands off of the steering wheel; and iii) cognitive distraction, where the driver takes their mind off the driving task, which can be due to singing, talking etc. The association of crashes with distracted behavior is underreported~\cite{dingus2016driver}. Verifying the distraction in the post-crash investigation is difficult, and the driver at fault might not disclose this information voluntarily. Among all the U.S. drivers involved in fatal
crashes in 2021, 5\% were reported as distracted at the time of the crashes~\cite{NHTSA2023}. Contrary to the numbers reported in crash analysis, Naturalistic Driving Data analysis by Dingus et al.~\cite{dingus2016driver} reported that almost 90\% of their studied crashes, which resulted in injury or property damage, can be attributed to driver-related factors (i.e.,  error, impairment, fatigue, and distraction). Numerous studies have explored the relation between precrash driver instability and distracted driving~\cite{klauer2006, arvin2019, arvin2020}. Arvin et al.~\cite{arvin2019} observed a substantial relation between distracted driving with crash intensity. They showed the probability of a severe crash increased by 11.1\% due to driver distractions. Therefore, real-time detection of the driver's status and providing effective feedback signals to minimize 
distraction is essential. It could significantly decrease the number of accidents resulting from distracted driving~\cite{young2007, kashevnik2021}. In recent years, with the rapid growth of affordable sensors and storage technologies, Naturalistic Driving Studies (NDS) have become the most common approach for analyzing driver behavior~\cite{Survey}. NDS offers video data which is preferred for investigating driver distractions in unbiased real-world settings. These studies leverage computer vision techniques to detect distracted driving, aiming to reduce crash risks by alerting drivers autonomously.
However, the reliance on human annotation of driver distraction in these studies introduces significant time delays. 
Our study aims to efficiently identify the common distraction activities in naturalistic driving videos using a vision-language-based fremework~\cite{clip} with limited annotated data, reducing training time and computational resources.

Recent advancements in deep learning and computer vision have significantly surged video analytics to identify distracted driving. In recent studies, the Convolutional Neural Networks (CNN) have been widely employed~\cite{dmd-dataset, dad, DHCNN, Faiqa2021, 3dcnn2019} for video-based distracted driving detection. However, these supervised CNN-based approaches require vast amounts of labeled data for training, making the process resource-intensive. This heavy reliance on extensive labeled data poses a significant challenge and can limit their generalization capabilities. Therefore, researchers have recently started to investigate unsupervised~\cite{unsupervised-Li} and semi-supervised~\cite{sam-dd-dataset} learning techniques.

Most of the distracted driving recognition tasks in computer vision primarily rely on analyzing in-vehicle video footage to detect driver inattentiveness~\cite{Survey,dad}. The challenge lies in the model's ability to handle diverse and unexpected driving behaviors that are common in real-world driving scenarios. Therefore, it is essential for the models to efficiently adapt and recognize a broader spectrum of actions, particularly focusing on those that are rare or challenging to classify. While traditional CNN models perform well on specific datasets, their success depends heavily on large-scale data and annotation efforts~\cite{unsupervised-Li}. Additionally, training of these models is often geared towards mapping predefined categories, which impedes their adaptability to new data. Therefore, rather than confining the learning to specific visual features within a single modality framework, learning from natural language supervision could provide adaptability and wide applicability~\cite{clip,videoclip, actionclip}. The recent advancements in vision-language pretraining frameworks showcase significant promise for diverse applications~\cite{dalle, flava,VLMo, VL-t5}. These frameworks utilize a multi-modal contrastive loss to generate robust embeddings that simultaneously address text and vision tasks. This approach allows several domains to use an almost zero-shot strategy for image-based object classification and detection tasks~\cite{clip, detic, actionclip, feuer2023zero}.

The study presented in this paper leverages CLIP's~\cite{clip} vision and language embeddings for identifying distracted driving actions. However, directly training such a model for video analysis is impractical due to the need for large-scale video-text training data and considerable GPU resources. An alternative and more feasible strategy is leveraging existing language-image models for application in the video domain~\cite{imgCLIP-to-vid,bottle-neck-video-text, coot}. Several recent studies have investigated strategies for transferring knowledge from these pretrained language-image models to other downstream tasks~\cite{ clipforge,jain2021dreamfields,clip2vid}.

Our proposed method leverages the pretrained CLIP encoders and offers two significant benefits. Firstly, the zero-shot transfer eliminates the requirement for extensive dataset-specific training, making it more scalable and practical for understanding video content through natural language. This contrasts with traditional methods that rely on memorizing human-annotated labels or features which are often non-transferable.  Secondly, it allows further finetuning of visual representations with a small amount of training data, thus enhancing the model's performance. Moreover, our approach addresses the evaluation issue encountered in previous studies~\cite{weiheng} by introducing cross-validation techniques and subject-level separation in training and testing sets.

\noindent In summary, the key contributions of this work include the following:
\begin{enumerate}
    \item We introduce and explore the performance of various vision-language-based frameworks (adapted from~\cite{clip}) for distracted driving action recognition tasks with minimum training. We present both frame-based and video-based models and show that including temporal information in the existing vision-language models offers superior performance on several distracted driving datasets. Moreover, we execute exhaustive fine-tuning with extensive hyperparameter tuning across multiple datasets to achieve the optimized performance. It tests the model's generalizability to previously unseen drivers. 
    
    \item We propose a unique validation method named `leaving a set of driver out', which involves subject-level separation during training to prevent overfitting to specific driver appearance. Moreover, we perform a cross-validation step to confirm the models' effectiveness in a variety of real-world conditions.
    
    \item We perform an in-depth analysis of the models' performance across different datasets and various experimental settings. Additionally, we demonstrate the model's robustness by maintaining high-performance levels even with reduced training data sizes. These attributes underscore the potential for deploying our models in real-world scenarios where training data may be limited.
\end{enumerate}

\noindent In this study, we develop and test four unique CLIP-based frameworks on four distinct distracted driving datasets. These frameworks include both frame-based and video-based models.

The rest of this paper is organized as follows: Section~\ref{sec:rel_work} provides a review of relevant research on distracted driving action recognition. Section~\ref{sec:method} includes the details of our methods. Section~\ref{sec:exp} provides details about the datasets and the experimental setup. Our results and observations are presented in Section~\ref{sec:result}. The paper concludes in Section~\ref{sec:conclusion}.

\section{Related Work}
\label{sec:rel_work}

In recent years, the area of driver action recognition and behavior analysis has attracted significant research interest. Generally, we notice two distinct categories of research works in the literature. 

\subsection{Visual Foundation methods}
Before the advent of computer vision techniques, distracted driving detection primarily relied on traditional machine learning methods, which involved manually selecting key features~\cite{body2, hand1,eye-gaze,drozy-data, eeg-LSTM} and pairing them with a simple classifier~\cite{ml-feature, HMM, random-forest, knn}.
However, the major limitation of these methods is that the manual feature selection lacks generalization, making it challenging to apply them effectively across diverse driving scenarios~\cite{Survey, toward}. However, the emergence of camera sensors and advancements in Convolutional Neural Networks (CNNs) have significantly influenced the field of driver behavior observation. Therefore, a significant portion of the current methodologies depends on various supervised CNN models~\cite{dmd-dataset,dad,DHCNN} and Deep Neural Networks (DNN)~\cite{FRNet, toward} for extracting important features from the image and video content. Some CNN-based frameworks gather features from facial recognition~\cite{facetracking}, optical flow~\cite{opticalflow}, and body posture assessment~\cite{skpose} for distraction action recognition. Some of the most common CNN backbones used in literature are VGG16~\cite{vgg16}, ResNet50~\cite{rn50}, AlexNet~\cite{AlexNet}, Inception-v3~\cite{incepv3}, MobileNet-v2~\cite{mobilenetv2}, ShuffleNet-v2~\cite{shufflenet} etc. However, the parameter size of CNN poses a challenge for practical applications~\cite{toward}. Therefore, some lightweight CNN frameworks~\cite{FRNet, mobilevgg} are proposed in the literature.
Various CNN approaches are employed based on the availability of dimensions (2D/3D), modalities (RGB, IR, Depth,~\cite{dad, dmd-dataset}), camera views (front, side etc.) of datasets and ensemble techniques~\cite{driveandact}. Traditional video-based action recognition CNN models often evolve from image-based 2D CNN models, adapting to incorporate the temporal aspect of videos. Strategies for integrating the temporal dimension include using 2D CNNs on multiple frames or 3D CNNs on video clips~\cite{quo}. While 3D CNNs can learn temporal features from video sequences, their implementation is often limited by high computational requirements and real-world challenges in processing naturalistic videos~\cite{3dcnndiba,D3d, quo}. Another technique that is commonly adopted involves ensembling multiple CNN networks~\cite{auc,hcf, ensemble}. Here, each CNN works as a unique feature extractor. Finally, they are incorporated into a weighted ensemble to yield a comprehensive global feature. In addition, there are methods that adopt feature-level fusion~\cite{quo,twostream1,twostream2}. These methods separately learn from RGB appearance and time-based features and later combine the learned features.

\subsection{Distracted Driving Recognition methods}
Before the widespread use of Vision Transformer (ViT) models~\cite{VIT}, some works relied on CNN for distracted driving detection. For instance, Moslemi et al.~\cite{3dcnn2019} proposed a 3D CNN model, which was pretrained on both the ImageNet
and Kinetics datasets and fine-tuned on a distracted driving detection dataset. This model was designed to learn spatiotemporal features from video sequences. Recently, the landscape of distracted action recognition is increasingly being shaped by ViT-based techniques~\cite{vivit, mvit,vidt,swin-T}, which have started to dominate video action recognition tasks. A ViT-based, weakly supervised contrastive (W-SupCon) learning approach is proposed in~\cite{sam-dd-dataset}, where it quantifies distracted behaviors by measuring their deviation from the normal driving representations. A key limitation of this study is the challenge of distinguishing high-risk driving behaviors from normal ones due to their similar physical movements indicating limited generalization.
A multilayer perceptron-based unsupervised model is proposed in~\cite{unsupervised-Li} where a ResNet50~\cite{rn50} backbone is trained for enhanced feature extraction. However, this approach's generalizability is limited, having been tested on only one dataset without large-scale validation. Moreover, a recent study by Chai et al.~\cite{weiheng} points out that many approaches in distracted driving detection research do not undergo thorough performance evaluation, leading to potentially inflated accuracy claims~\cite{Survey, dhakate2020, JiewrongAcc2021}. They highlight the significance of implementing driver-level separation and cross-validation techniques during evaluation to address these shortcomings.

The recent progress in vision-language pretraining approaches~\cite{videoclip, actionclip, movieCLIP} has demonstrated improved performance in tasks related to video action recognition. These models are developed on a widely used vision-language model named CLIP~\cite{clip}.
For video-based CLIP training, Miech et al.~\cite{miech2019howto100m} introduce a large dataset referred to as HowTo100M. Additionally, for learning the visual-semantic joint embedding, Mithun et al.~\cite{mithun2018webly} propose a two-stage approach that leverages web images and corresponding tags. Dong et al. ~\cite{dong2019dual} attempt to solve the challenging problem of zero example video retrieval using textual features. Therefore, various methods have been developed to effectively learn from the textual and visual-semantic joint embeddings. These methods underline the potential of integrating visual and textual data in developing advanced multimodal models.

\section{Methodology}
\label{sec:method}
\begin{figure*}[tbh]
\centering
\includegraphics[width=5in]{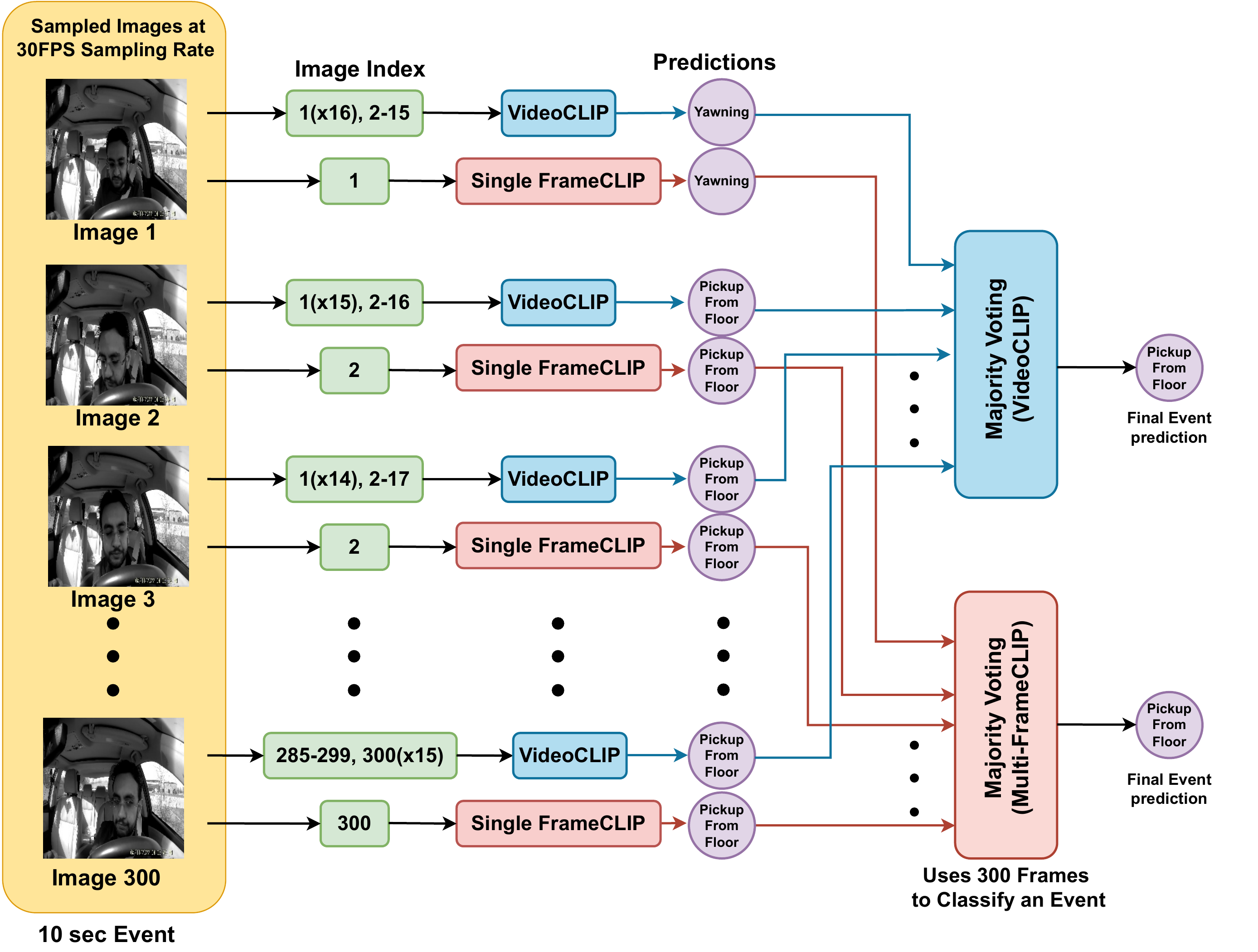}
\caption{An overview of the image-based and video-based CLIP frameworks. The yellow box shows the sampled frames from a 10-second event, i.e., a total of 300 frames at a rate of 30 frames per second. These frames are sampled from a raw video. The VideoCLIP uses 30 consecutive frames to create a short video clip as input, while the Single-frameCLIP takes a single frame at a time. For the very first sample, the VideoCLIP generates a video clip by repeating Image-1 sixteen times and including Images 2 to 15 (a total of 30 frames in the green box). The SingleframeCLIP only uses Image-1 as its first input and makes the prediction. This process continues for each subsequent frame.  Note that the prediction for an entire event is made by calculating the majority vote from all individual frame predictions. The Multi-frameCLIP applies majority voting on SingleframeCLIP's predictions and VideoCLIP applies majority voting on its output to get the final event prediction.}
\label{modelpipeline}
\end{figure*}
This study aims to understand and analyze distracted driving behaviors from in-vehicle video footage and build an effective driver monitoring system. Despite the wide range of driving distractions, we only focus on specific actions or events that divert the driver's attention from safe driving. Furthermore, we consider each unique distracted action that lasts for a certain duration as an individual event. We also assume that these distracting events do not overlap or occur simultaneously. For example, distracting events can be talking on the phone, reaching behind, or drinking water while driving. Detecting these distracting events from video data involves several challenges, including low-quality video and the extensive human effort required to annotate videos or images for training supervised models. Therefore, we aim to develop a generalized framework that performs well without strictly relying on labeled data or with only limited labeled data. Our proposed framework is based on a contrastive vision-language pretraining approach designed to address these challenges with minimal annotated data. Moreover, we carefully conduct our evaluation experiments to ensure a robust and generalizable testing strategy. In real-world scenarios, the model will be expected to encounter new and unseen distracting events. Therefore, we employ two methods during training: 1) subject-wise separation method -~\textit{‘leaving a set of driver out’}  where we separate out a unique set of driver samples only for testing. 2) subject-wise cross-validation method -~\textit{$k$-fold cross-validation}  where we implement $k$-fold cross-validation technique individually to each driver set. For the first method, the ``test split drivers' data" are never presented to the model during its training phase. It ensures that the model only focuses on distracting events rather than the subject's appearance and can make predictions fully on unseen data during testing. These two methods ensure zero overlapping between training and testing splits and zero memorizing across all the runs we conducted. The average accuracy obtained from all the $k$-fold results is reported as the final accuracy of our frameworks. This process allows us to accurately assess the overall performance of our proposed models across all the datasets.

Furthermore, we recognize the significance of addressing multiple distracting events occurring simultaneously or alternately. Our observations have revealed that the effectiveness of handling such scenarios depends on the quality of the datasets and the formulation of the prompts. It is essential to underscore that dealing with multiple distracted behaviors is indeed a complex challenge within the scope of this problem. While our current framework has not been specifically tested on multiple distracting events, we firmly believe that, with the availability of appropriate datasets and prompt customization, our frameworks can be adapted to detect and analyze such situations. We acknowledge that this aspect of the problem requires further research and exploration.

\subsection{Proposed Frameworks}
We fundamentally propose two types of models: (i) frame-based models (Zero-shotCLIP, Single-frameCLIP, and Multi-frameCLIP) and (ii) video-based models (VideoCLIP). An overview of these frameworks is provided in Fig.~\ref{modelpipeline}. CLIP~\cite{clip} is a popular vision-language model. It is proficient in most visual understanding tasks since it was trained on a large variety (around 400 million) of image-text pairs from the internet. \\
\begin{figure}[b!]
\centering
\includegraphics[width=3.5in]{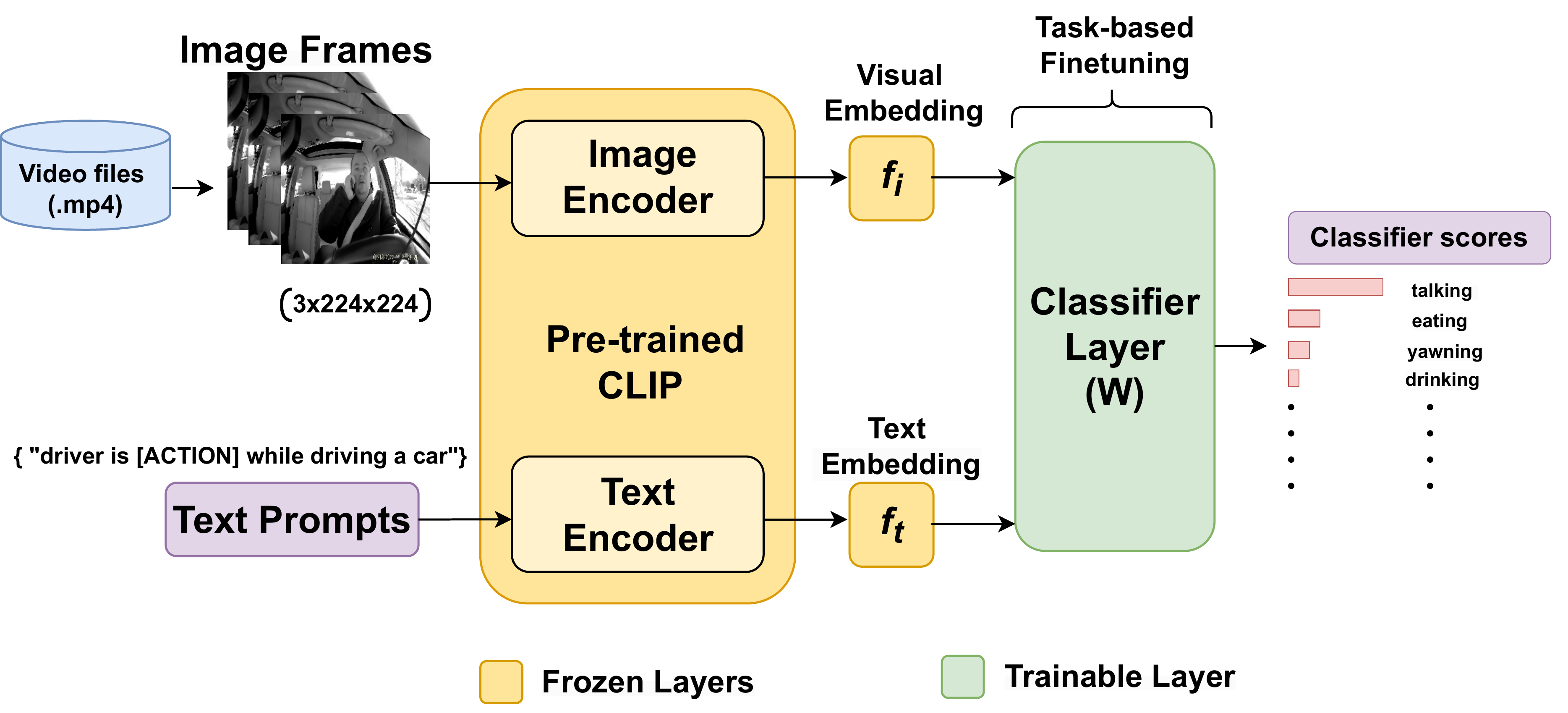}
\caption{The proposed multimodal Single-frameCLIP framework exploits the pretrained CLIP architecture to extract vision-text semantic information.}
\label{fig:clip}
\end{figure}

\subsubsection{Frame-based models}
The frame-based models operate by taking inputs similar to the original CLIP model and making predictions on distracted actions. As shown in Fig.~\ref{fig:clip}, the model projects the input image and text pairs into a multi-modal latent space and minimizes a contrastive loss, which enables it to learn meaningful and useful visual and textual representations. Therefore, the base model can be fine-tuned for task-specific learning. Given its ability to be instructed by natural language for a broad range of downstream tasks, such as video question answering and understanding actions from video, we develop our distracted driving analysis framework based on this model. A great advantage of using a pre-trained network over other strategies is the ability to utilize the pre-existing visual representation of the language prompt, offering the zero-shot transfer. In this work, we initially leverage CLIP's pre-trained visual embedding for zero-shot transfer on distracted driving datasets and categorize the distracted action classes without any additional training. Following this, we fine-tune a linear classifier layer built upon the original pre-trained CLIP model. The yellow block shown in Fig.~\ref{fig:clip} represents the frozen layers. We add the green block positioned on top of the yellow layers, which represents the trainable layer. This task-based fine-tuning green block involves training a straightforward classifier layer using the pre-existing embeddings from the frozen layers for a specific task. In our case, this task is to classify distracted driving actions from visual data. We evaluate the performance of the whole framework on corresponding distracted driving datasets. As per the primary model~\cite{clip}, we implement five CLIP visual backbones- ViT-L/14, ViT-L/14@336px, ViT-B/16, ViT-B/32, RN101 and assess their performance individually.

The Zero-shotCLIP, Single-frameCLIP and Multi-frameCLIP frameworks have input $I\in \mathbb{R}^{3 \times H \times W}$ where $I$, $H$, and $W$ represent the input image, image height and image width, respectively. The Zero-shotCLIP makes predictions on raw frames without requiring any training. The key difference between the other two frameworks is that Single-frameCLIP makes predictions looking at a single frame and the Multi-frameCLIP makes predictions on a stack of frames conducting a majority vote at the classification layer. For both frame-based and video-based CLIP, all the layers up to the classifier layer are kept frozen and we only finetune the last trainable layer (the green layer in Fig.~\ref{fig:clip}).

\subsubsection{Video-based model}



The VideoCLIP model has a similar structure to the frame-based model shown in Fig.~\ref{fig:clip}, but uses a different CLIP visual encoder and model input. It utilizes a branch called S3D (separable 3D CNN)~\cite{3dcnn}, a video processing backbone based on 3D-CNN, which takes a video $V \in \mathbb{R}^{S\times 3 \times H \times W}$ as input and encodes the video into a feature vector $F$. Here $S$, $V$, $H$, and $W$ represent the number of frames, image height and image width, respectively. In our experiments, the number of frames $S$ is set as 30, indicating that the VideoCLIP model uses 30 frames and synthesizes them into a short video clip. The rest of the training process is similar to the frame-based CLIP. In the training process, if the cosine similarity between the video feature and its corresponding text feature is high, the distance between their feature vectors is minimized. It should be highlighted that both the Multi-frameCLIP and VideoCLIP operate on successive frames, thereby capturing the temporal context of a given distracted action. A majority vote strategy is employed at the final stage by both frameworks. The key difference between these two models is illustrated in Fig.~\ref{modelpipeline}.

\section{Experiments}
\label{sec:exp}
\subsection{Dataset} 
In this study, we utilize four common distracted driving datasets—DMD~\cite{dmd-dataset}, SAM-DD~\cite{sam-dd-dataset}, SynDD1~\cite{syndd1} and StateFarm~\cite{statefarm} to investigate driver behavior under various real-world conditions. These datasets (summarized in Table~\ref{tab:datasets}) serve as the foundation for evaluating our proposed frameworks, offering a wide variety of annotated images and videos with diverse distracted driving scenarios.  The action categories span common distractions that affect driving safety, providing a comprehensive resource for analyzing distracted driving behavior. In addition, they incorporate a range of modalities including RGB, IR, and Depth, with input resolutions up to 1920x1080. The datasets used in our study are specifically chosen for their unique driver IDs for each participant. It is crucial to conduct the `leave a set of driver out' evaluation.

The usage of these diverse datasets ensures the generalizability of our proposed framework by incorporating a wide range of distraction actions, data modalities, input resolutions, and various camera views to capture comprehensive driving behaviors across different scenarios and conditions. A significant number of drivers (ranging from 18 to 42) with varied driving habits and physical characteristics further enriches the data pool, enabling our framework to learn from a broad spectrum of real-world driving situations. This diversity allows us to develop robust and adaptable frameworks capable of accurately identifying distracted driving behaviors across different vehicles, environments, and driver demographics. As a result, our frameworks can be optimized to perform effectively in real-world applications, reducing the risk of overfitting to specific conditions and enhancing their applicability across a wide range of driving contexts.

\begin{table*}[htb]
\begin{center}
\setlength{\abovecaptionskip}{2pt}
\caption{Overview of Distracted Driving Datasets}
\label{tab:datasets}
\setlength{\tabcolsep}{5pt}
\begin{tabular}{*{8}{c}}
\toprule
Dataset & Modality  & Input& Data& No. of & No. of & No. of  & Action \\
name & & resolution&type & camera & distraction & drivers & categories \\
 & & & &view &actions & & \\
\midrule DMD~\cite{dmd-dataset} & \begin{tabular}{@{}c@{}@{}}RGB \\ IR\\ Depth\end{tabular} & 1920x1080&\begin{tabular}{@{}c@{}}Image \\ Video\end{tabular} & 2 & 10 & 37 & \begin{tabular}{@{}c@{}@{}@{}}drinking, hair and makeup, talking on phone (right)\\ operating radio, reaching behind, reaching side, safe driving\\talking to passenger, texting (right), yawning\end{tabular}\\
\midrule SAM-DD~\cite{sam-dd-dataset} & RGB & 1200x900& \begin{tabular}{@{}c@{}}Image \\ Video\end{tabular} & 2 & 10 & 42 & \begin{tabular}{@{}c@{}@{}@{}}safe driving, drinking, talking on phone (left),\\talking on phone (right), texting (left), texting (right),\\hair, adjusting glasses, reaching behind, head dropping\end{tabular}\\
\midrule StateFarm~\cite{statefarm} & RGB&640x480& Image & 1 & 10 & 26 & \begin{tabular}{@{}c@{}@{}@{}}safe driving, texting (right), talking on phone (right),\\ texting (left), talking on phone (left), operating radio, drinking, \\reaching behind, hair and makeup, talking to passenger\end{tabular}\\
\midrule SynDD1~\cite{syndd1} & RGB &1920x1080& \begin{tabular}{@{}c@{}}Image \\ Video\end{tabular} & 3 & 8 & 18 & \begin{tabular}{@{}c@{}@{}}adjusting hair, drinking, eating,\\ picking up something, reaching behind,\\  singing, talking on phone, yawning\end{tabular}\\
\bottomrule
\end{tabular}
\end{center}
\end{table*}

\begin{table*}[htb]
\begin{center}
\setlength{\abovecaptionskip}{2pt}
\caption{Summary of Experimental setup}
\label{tab:exp_setup}
\setlength{\tabcolsep}{5pt}
\begin{tabular}{*{7}{c}}
\toprule
Framework & Approach  & Input& Training& Cross & Leaving a set  \\
name & & resolution&details & validation &of driver out \\
\midrule Zero-shotCLIP & \begin{tabular}{@{}c@{}}Zero-shot\\ transfer\end{tabular} & 224x224x3 &\begin{tabular}{@{}c@{}}No training, \\ use pre-trained CLIP\end{tabular} & N/A& N/A\\
\midrule Single-frameCLIP & \begin{tabular}{@{}c@{}} Fine-tuning\\ classifier-layer\end{tabular} & 224x224x3 & \begin{tabular}{@{}c@{}}Training linear classifier layer \\ batch size: 100, optimizer: LBFGS\end{tabular} & \begin{tabular}{@{}c@{}} 8-fold (StateFarm), \\ 7-fold (others)\end{tabular} &\checkmark\\
\midrule Multi-frameCLIP & \begin{tabular}{@{}c@{}} Sequence-analysis\\ with majority voting\end{tabular} &variable& \begin{tabular}{@{}c@{}}Single-frameCLIP predictions \\ aggregated with majority voting\end{tabular} & 7-fold &\checkmark\\
\midrule VideoCLIP & \begin{tabular}{@{}c@{}} Video feature-analysis\\ with temporal context\end{tabular}  &\begin{tabular}{@{}c@{}} 224x224x3x30\\ (30 frames)\end{tabular} & \begin{tabular}{@{}c@{}}Learning rate: 0.0003 \\ batch size: 5, optimizer: Adam\end{tabular} & 7-fold&\checkmark\\
\bottomrule
\end{tabular}
\end{center}
\end{table*}

\subsection{Experiment Setup}  
Table~\ref{tab:exp_setup} summarizes our experimental setup across the image-based and video-based models. It shows how we leverage the versatility of the CLIP model to learn distracted driving patterns, evolving from zero-shot learning to advanced video analysis. First, we combine the pre-trained CLIP backbones with sentence-level prompts in Zero-shotCLIP to get predictions on raw images without prior training. Then, the Single-frameCLIP extends this idea by fine-tuning an additional classifier layer on our distracted driving datasets. Finally, Multi-frameCLIP applies a majority voting on top of Single-frameCLIP logits and VideoCLIP incorporates a 3D CNN backbone~\cite{3dcnn} on consecutive frames to make a ``distraction event" prediction. We maintain the standard CLIP input resolution shown in the second column of the table.
The original CLIP model fine-tuning requires sentence descriptions as labels rather than the traditional single text label. Therefore, we provide a full sentence description as a prompt for each action category. As reported in~\cite{clip}, we consider manually fine-tuning and rephrasing the prompts for our experiments, an area that can be further optimized in future research. While testing various hyperparameter settings (backbones, FPS, camera-view etc.),  we select the parameters that deliver the highest performance for our proposed models and refer to them as the ``optimal hyperparameter setting". Furthermore, to compare the performance of our frameworks with traditional baselines, we trained five CNN backbones  (AlexNet~\cite{AlexNet}, VGG16~\cite{vgg16}, MobileNet-v2~\cite{mobilenetv2}, ResNet18 and ResNet50~\cite{rn50}) on the same dataset. For this experiment, we load the ImageNet pre-trained weights to train these models with a learning rate of 0.001, a batch size of 64 and Adam~\cite{kingma2014adam} as the optimizer.  Moreover, to resolve the issue raised in~\cite{weiheng}, we design the experiments to leave a specific subset of drivers separate only for testing rounds, ensuring a strict separation between training and testing splits. In addition, we adopt a $k$-fold cross-validation method on each driver set, where the final model accuracy is assessed by averaging accuracies over all folds. Furthermore, to maintain consistency and reliability in our experimental assessments, we maintain uniform experimental settings across all models and datasets. The results of these experiments along with the effect of the optimal hyperparameter setting are presented in Section~\ref{sec:result}.

\section{Results}
\label{sec:result}
\subsection{Performance summary of the CLIP-based Frameworks}
\begin{table}[htb]
\centering
\setlength{\abovecaptionskip}{2pt}
\caption{Performance Summary of CLIP-based \\Frameworks on Distracted driving datasets}
\begin{tabular}{lcccccc}
\toprule
\textbf{Models} &\textbf{SynDD1} &\textbf{StateFarm}&\textbf{DMD}& \textbf{SAM-DD}\\
\textbf{(ours)}&Top-1& Top-1&Top-1& Top-1\\
\midrule
Zero-shotCLIP & 54.00& 58.22 & 48.12 & 66.52 \\
Single-frameCLIP & 57.35& \textbf{83.15}& 82.65 & 89.14\\
Multi-frameCLIP & 70.36 & N/A &93.39 & 89.14 \\
VideoCLIP & \textbf{81.85} & N/A &\textbf{98.44} & \textbf{97.86} \\
\bottomrule
\end{tabular}
\label{tab:clip_overall}
\end{table}

We present the performance summary of various CLIP-based frameworks on the four distracted driving datasets in Table~\ref{tab:clip_overall}. 
The VideoCLIP model with an optimal hyperparameter setting demonstrated the most superior performance with a Top-1 accuracy of 81.85\%, 98.44\% and 97.86\% on SynDD1, DMD and SAM-DD datasets, respectively. The multi-frameCLIP framework also showed notable results with its optimal hyperparameter setting. Single-frameCLIP and Zero-shotCLIP reported lower Top-1 accuracies on the SynDD1 dataset compared to the SAM-DD dataset due to the dataset size. On the StateFarm dataset, only Single-frameCLIP and Zero-shotCLIP were evaluated due to the absence of video data, which is a prerequisite for Multi-frameCLIP and VideoCLIP. Single-frameCLIP 
outperformed Zero-shotCLIP with optimal hyperparameter setting, achieving a Top-1 accuracy of 83.15\%. These results highlight the effectiveness of VideoCLIP and Multi-frameCLIP in leveraging temporal information from video data for improved performance. 



\subsection{Performance of the Frameworks with Optimal Hyperparameter Setting}

\begin{table*}[htb]
\begin{center}
\setlength{\abovecaptionskip}{2pt}
\caption{Proposed frameworks performance  with Optimal Hyperparameter setting}
\label{tab:CLIPresults-optimal-hyp}
\setlength{\tabcolsep}{5pt}
\begin{tabular}{*{7}{c}}
\toprule
Framework & Dataset & CLIP& Camera& FPS &Top-1 & Leaving a set\\
name & & backbone& view&&accuracy& of driver out \\
\midrule 
\multirow{4}{*}{Zero-shotCLIP}& DMD& ViT-L/14@336px& Side-view &N/A&48.12& N/A \\
& SAM-DD& ViT-L/14@336px & Side-view &N/A&66.52&N/A \\
& StateFarm& ViT-L/14@336px & Side-view &N/A&58.22 &N/A \\
& SynDD1& ViT-L/14@336px & Dashboard &N/A&54.00&N/A\\
\midrule 
\multirow{4}{*}{Single-frameCLIP} & 
DMD& ViT-L/14@336px& Side-view &N/A&82.65 & \checkmark \\
& SAM-DD& ViT-L/14 & Side-view &N/A& 89.14 & \checkmark \\
& StateFarm& ViT-L/14 & Side-view &N/A&83.15 & \checkmark \\
& SynDD1& ViT-L/14 & Dashboard &N/A&57.35 & \checkmark \\
\midrule 
\multirow{3}{*}{Multi-frameCLIP} & DMD& ViT-L/14@336px& Side-view &1&93.39 & \checkmark \\
& SAM-DD& ViT-L/14 & Side-view &1&89.14 & \checkmark \\
& SynDD1& ViT-L/14 & Dashboard &1&70.36 & \checkmark \\
\midrule 
\multirow{3}{*}{VideoCLIP} & DMD& ViT-L/14@336px& Side-view &1&\textbf{98.44} & \checkmark \\
& SAM-DD& ViT-L/14 & Side-view &1&\textbf{97.86} & \checkmark \\
& SynDD1& ViT-L/14 & Dashboard &20&\textbf{81.85} & \checkmark \\
\bottomrule
\end{tabular}
\end{center}
\end{table*}

Table~\ref{tab:CLIPresults-optimal-hyp} summarizes the performance of Zero-shotCLIP, Single-frameCLIP, Multi-frameCLIP and VideoCLIP—across various datasets with optimal hyperparameter settings. Each framework achieves its best results using either the ViT-L/14 or the ViT-L/14@336px backbone, evaluated on specific datasets. Performance is measured in terms of Top-1 accuracy, with all experiments incorporating a ``leaving a set of driver out" and a $k$-fold cross-validation step. Zero-shotCLIP serves as a benchmark for initial model results. Single-frameCLIP demonstrates consistent performance with accuracies ranging from 57.35\% to 89.14\% across different views and datasets. Multi-frameCLIP exhibits performance enhancements, highlighting a significant accuracy increase to 93.39\% on the DMD dataset with a side-view camera setting. VideoCLIP surpasses the other models, achieving peak accuracy scores on all three datasets. It demonstrates the significant advantage of leveraging temporal features and video-based feature representation. This result highlights the effectiveness of adopting the VideoCLIP model and the substantial improvements possible by incorporating video-based features over frame-based methods. It also illustrates how the selection of optimal hyperparameters such as camera views, sampling rates, and CLIP backbones, plays a pivotal role in maximizing model performance.

\subsection{Performance Comparison with Other Works}
In this section, we present a performance comparison of our proposed CLIP-based frameworks (both frame-based and video-based) with traditional distracted driving recognition models.

\subsubsection{Single-frameCLIP vs. Traditional Models}
In this section, we compare the performance of the Single-frameCLIP model with the traditional distracted driving models on frame-based datasets. 
\begin{table}[ht!]
\begin{center}
\setlength{\abovecaptionskip}{2pt}
\caption{Single-frameCLIP vs. traditional models}
\label{tab:single-frameCLIP-vs-other}
\begin{tabular}{*{4}{c}}
\toprule 
\multirow{2}{*}{Dataset} & \multirow{2}{*}{Models} & Top-1 & Leaving a set \\
& & accuracy& of driver out \\
\midrule 
\multirow{4}{*}{StateFarm} & Single-frameCLIP & \textbf{83.15} & \checkmark \\
& ResNet18 & 82.68 & \checkmark \\
& MobileNet-v2 & 80.04 & \checkmark \\
& AlexNet & 74.84 & \checkmark \\
\midrule 
\multirow{4}{*}{SynDD1} & Single-frameCLIP & \textbf{57.35} & \checkmark \\
& ResNet18 & 42.62 & \checkmark \\
& MobileNet-v2 & 36.10 & \checkmark \\
& AlexNet & 13.71 & \checkmark \\
\bottomrule
\end{tabular}
\end{center}
\end{table}

\begin{figure}[ht]
\centering
\includegraphics[width=2.8 in, height= 2.5in]{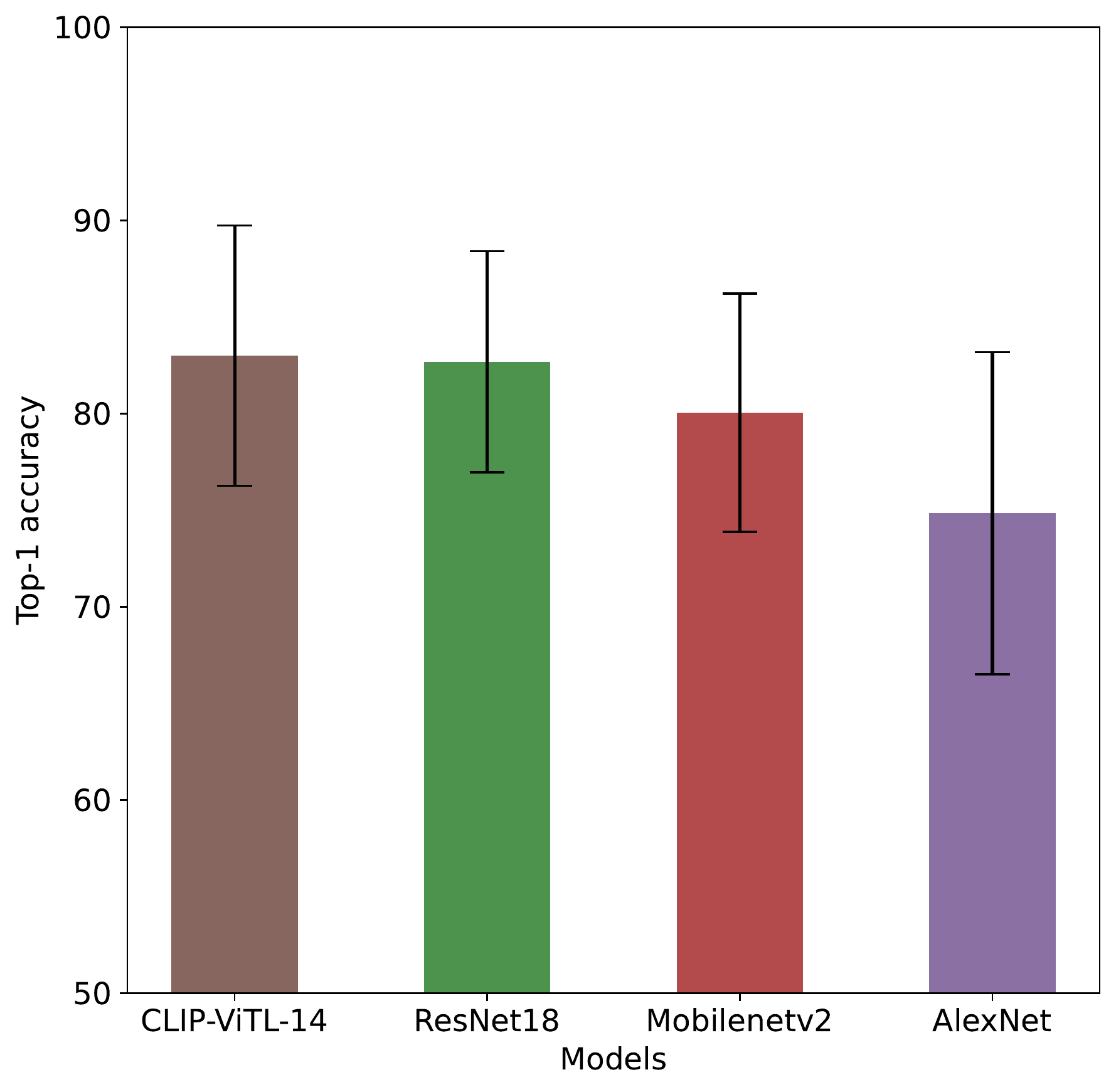}
\caption{ The mean and variance of Top-1 accuracy of the Single-frameCLIP and CNN baselines on the StateFarm dataset.}
\label{fig:single-frameCLIPvstradmodel}
\end{figure}

Table~\ref{tab:single-frameCLIP-vs-other} shows the performance of the Single-frameCLIP model against traditional Convolutional Neural Network-based (CNN) baselines on two frame-based datasets (StateFarm and SynDD1) in terms of Top-1 accuracy. The Single-frameCLIP based on ViT-L/14 backbone outperforms the traditional CNN models on both datasets. These results suggest that the Single-frameCLIP is more effective at leveraging contextual information and understanding complex visual patterns compared to traditional CNNs. It also demonstrates that the Single-frameCLIP generalizes better to unseen drivers (as indicated by the ``leaving a set of driver out" column) and diverse driving scenarios, making it a more effective framework for analyzing driver behavior.
Moreover, Fig.~\ref{fig:single-frameCLIPvstradmodel} represents the mean and variance of Top-1 accuracy of the Single-frameCLIP and CNN baseline models on the StateFarm dataset. The bars illustrate the average Top-1 accuracy of the models, while the vertical lines on the top indicate the variance in Top-1 accuracies across all the folds. This barplot highlights the consistent performance of the Single-frameCLIP model in distracted driving detection.


\subsubsection{Dataset size vs. Performance}
In an effort to analyze the performance of the Single-frame CLIP model with smaller amounts of training data, a series of experiments are conducted utilizing progressively reduced training datasets. In this experiment, we train the traditional CNN models and our proposed Single-frameCLIP model with a smaller training data sampled from the StateFarm dataset. We separate 3 drivers (driverID p012, p014, p021) for testing and progressively removed four drivers from the training set to create a smaller training set, which is $80\%$, $60\%$, $40\%$, and $20\%$ of the original training set. In this procedure, rather than indiscriminately eliminating random frames, we selectively remove the training samples according to driverID. This approach mirrors real-world scenarios where the model must adapt its learned behaviors to new and unseen drivers. The reduced training data, number of drivers present in the training splits, and the number of samples are shown in Table~\ref{tab:DriverTrainingSamples}.

\begin{table}[ht]
\begin{center}
\setlength{\abovecaptionskip}{2pt}
\caption{Training Data for the reduced training set Experiment}
\label{tab:DriverTrainingSamples}
\begin{tabular}{*{4}{c}}
\toprule
Percentage of & No. of drivers  & No.of samples& Test set\\
training data & in training &  in training&driver IDs\\
\midrule 80\% & 18 & 16163 & p012p014p021\\
\midrule 60\% & 14 & 12900&p012p014p021 \\
\midrule 40\% & 9 & 8949&p012p014p021 \\
\midrule 20\% & 5 & 5649&p012p014p021 \\
\bottomrule
\end{tabular}
\end{center}
\end{table}

\begin{figure}[htb]
\centering
\includegraphics[width=3.2in, height= 2.5 in]{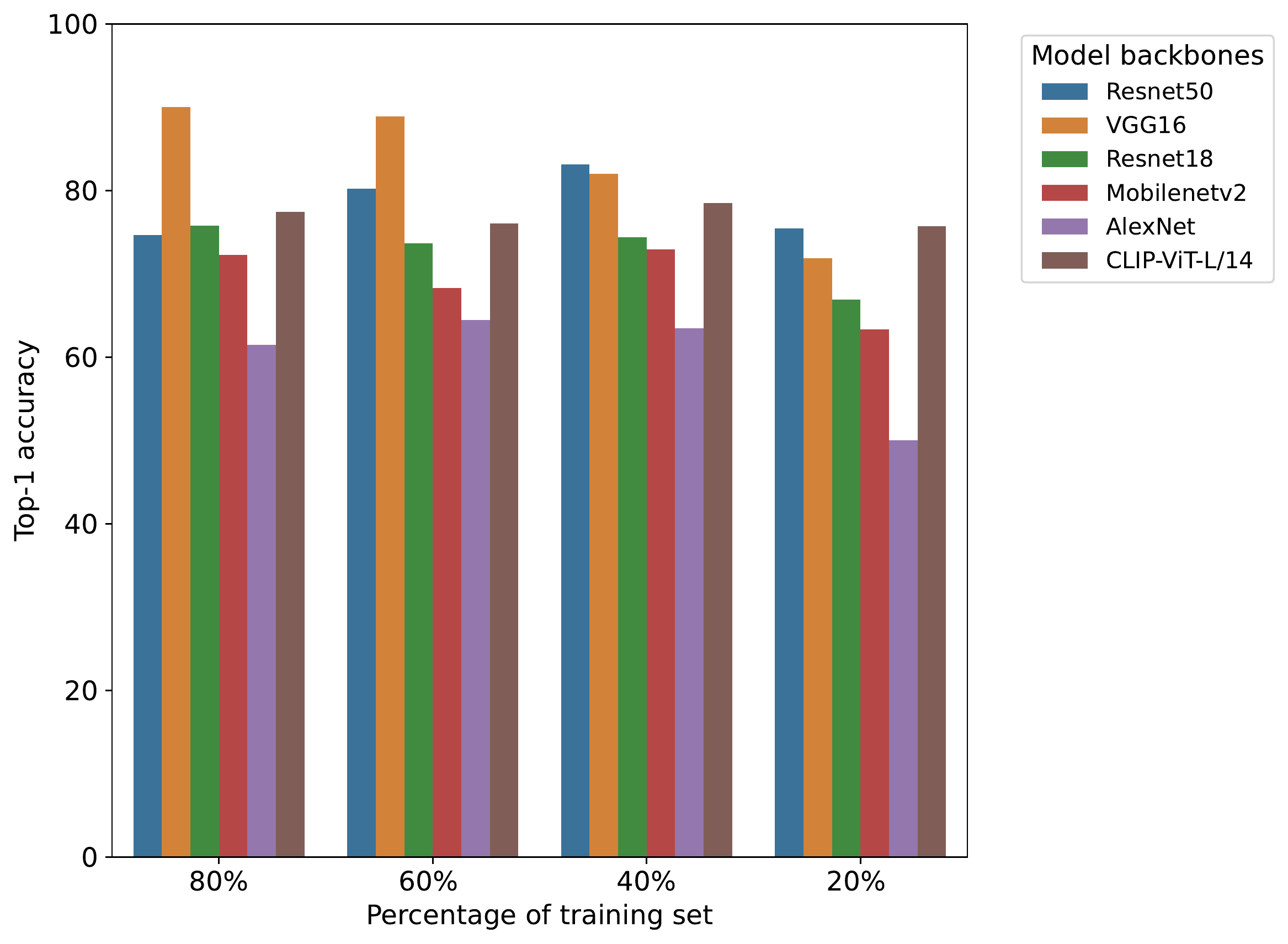}
\caption{Performance comparison of the Single-frameCLIP model and traditional CNN models trained on varying proportions of the StateFarm dataset. It shows average accuracies for 8-fold cross-validation.}
\label{redtrainSF}
\end{figure}

For this experiment, we employ 8-fold cross-validation and excluded three drivers (driverIDs are on the rightmost column) from every fold and used them as test subjects. We utilize the CLIP ViT/L-14 as the backbone of Single-frameCLIP. All the models use the same training samples and are finally tested on the same test set. The results in Fig.~\ref{redtrainSF} suggest that reducing the size of the training set does not necessarily compromise the Single-frameCLIP model's performance. Unlike the other models, which displayed a decrease in accuracy, the Single-frameCLIP model maintained its performance levels consistently even when the quantity of training data was reduced to 20\% of the original size.

\subsubsection{Multi-frameCLIP vs. Traditional Models}
\begin{table}[ht]
\setlength{\abovecaptionskip}{-5pt}
\begin{center}
\caption{Multi-frameCLIP vs. traditional models\\
(at Sampling rate 1FPS)}
\label{tab:Multi-frameCLIPvsOthermodel}
\resizebox{\columnwidth}{!}{%
\begin{tabular}{*{5}{c}}
\toprule  Video& Models & Top-1& Modality&Leaving a set\\ Datasets& & Acc.(\%) & &of driver out\\
\midrule \multirow{5}{*}{DMD} & AlexNet & 45.86 &RGB&\checkmark \\
& VGG16 & 56.15&RGB&\checkmark  \\
& MobileNet-v2& 85.77&RGB&\checkmark \\
& ResNet18& 85.95&RGB&\checkmark \\
& Multi-frameCLIP&\textbf{93.39}&RGB&\checkmark\\
\midrule \multirow{4}{*}{SAM-DD} & VGG16 & 86.93&RGB&\checkmark  \\
& AlexNet & 88.50&RGB&\checkmark  \\
& MobileNet-v2&\textbf{92.95}&RGB&\checkmark \\
& Multi-frameCLIP&89.14&RGB&\checkmark\\
\midrule \multirow{6}{*}{SynDD1} & AlexNet & 12.56 &RGB&\checkmark \\
& VGG16 & 12.56&RGB&\checkmark  \\
& MobileNet-v2& 33.65&RGB&\checkmark \\
& ResNet50& 43.49&RGB&\checkmark \\
& ResNet18& 47.55&RGB&\checkmark \\
& Multi-frameCLIP&\textbf{70.36}&RGB&\checkmark\\
\bottomrule
\end{tabular}}
\end{center}
\end{table}

We provide an extensive performance comparison of Multi-frameCLIP with traditional models on three video datasets in Table~\ref{tab:Multi-frameCLIPvsOthermodel}. To ensure consistency and reliability in our comparison, we run the listed models at a 1FPS sampling rate with the same 7-fold cross-validation and ``leaving a set of driver out" settings across all the datasets. The Multi-frameCLIP model shows superior performance compared to the recent baseline models, achieving a Top-1 accuracy of 93.39\% on DMD, 89.14\% on SAM-DD, and 70.36\% on the SynDD1 dataset. These results are significantly higher than the recent baselines when tested under the same setting. In addition, it showcases the Multi-frameCLIP model's efficacy in capturing driver behavior in different scenes, making it a suitable benchmark. However, it is important to note that accuracy fluctuates with changes in hyperparameters such as model type, sampling rate, camera view, and so on, indicating that these are significant factors to consider to get an optimized performance.

\subsubsection{VideoCLIP vs. Traditional Models}

\begin{table}[ht]
\setlength{\abovecaptionskip}{-5pt}
\begin{center}
\caption{VideoCLIP vs. traditional models\\(at Sampling rate 1FPS)}
\label{tab:VideoCLIPvsOthermodel}
\resizebox{\columnwidth}{!}{%
\begin{tabular}{*{5}{c}}
\toprule  Video& Models & Top-1& Modality&Leaving a set\\ Datasets& & Acc.(\%) & &of driver out\\
\midrule \multirow{9}{*}{DMD} & ShuffleNet-v2 & 84.80 &RGB& x \\
& 3D-ShuffleNet-v2& 88.50&RGB& x \\
& MobileNet-v2& 89.90&RGB&x\\
& 3D-MobileNet-v2& 90.30&RGB&x\\
& MobileNet-v2(MLP)& 91.60&RGB&x\\
& Inception-v3& 91.80&RGB& x \\
& MobileNet-v2(MLP)& 93.70&RGB+IR+D&x\\
& MobileNet-v2(MLP)~\cite{dmd-dataset}& 93.90*&IR+D&x\\
& VideoCLIP &\textbf{98.44}&RGB&\checkmark\\
\midrule \multirow{6}{*}{SAM-DD} & MobileNet-v3&94.78&RGB&\checkmark\\
& ShuffleNet-v2 & 95.56 &RGB&\checkmark\\
& ResNet18 & 95.91 &RGB&\checkmark\\
& VGG16 & 96.86 &RGB&\checkmark\\
& ViT & 97.09 &RGB&\checkmark\\
& VideoCLIP&\textbf{97.86}&RGB&\checkmark\\
\midrule SynDD1 &VideoMAE~\cite{aicity23top}&71.35*&RGB&\checkmark \\
& VideoCLIP(1FPS)& 74.64&RGB&\checkmark\\
& VideoCLIP(20FPS)& \textbf{81.85}&RGB&\checkmark\\
\bottomrule
\end{tabular}}
\footnotesize
\begin{flushleft}
*average Top-1 accuracy for the same classes used in VideoCLIP
\end{flushleft}
\end{center}
\end{table}

Table~\ref{tab:VideoCLIPvsOthermodel} showcases a comparative analysis of the VideoCLIP against the recent baseline models. It shows that VideoCLIP gets significant performance gain and outperforms its counterparts across three different datasets. For example, in the DMD dataset, VideoCLIP outperforms the other models, including MobileNet-v2 (MLP), and 3D CNN variants, achieving a Top-1 accuracy of 98.44\%. Similarly, in the SAM-DD dataset, VideoCLIP tops the chart with 97.86\% accuracy, surpassing other state-of-the-art models such as ViT and VGG16. These results highlight the efficacy of the VideoCLIP model in harnessing the RGB modality of the input video and underscore its robustness across different video datasets.

\begin{table}[htb]
\begin{center}
\setlength{\abovecaptionskip}{2pt}
\caption{Single-frameCLIP: F1-score on DMD and SAM-DD}
\label{class-level}
\begin{tabular}{ccc}
\toprule
Action & F1-score&F1-score\\
class & DMD &SAM-DD\\
\midrule drinking &0.96&0.98\\
\midrule hair and makeup &0.91&0.84\\
\midrule talking to phone& 0.97&0.96 \\
\midrule adjusting radio &0.86&-\\
\midrule reaching to backseat&0.68&0.91 \\
\midrule reaching to side &0.58&- \\
\midrule driving safely&0.68&0.93\\
\midrule talking to passenger &0.87&- \\
\midrule texting on phone&0.92&0.84 \\
\midrule yawning &0.76&- \\
\bottomrule
\end{tabular}
\end{center}
\end{table}

\subsection{Hyperparameters and Ablation Studies}
In this section, we discuss how various real-world parameter settings, such as the sampling rate, camera location and presence of different distracted driving classes can impact the performance of our proposed CLIP-based frameworks.

\subsubsection{Class-level Analysis}

\begin{figure}[htb]
\centering
\includegraphics[width=3.2in]{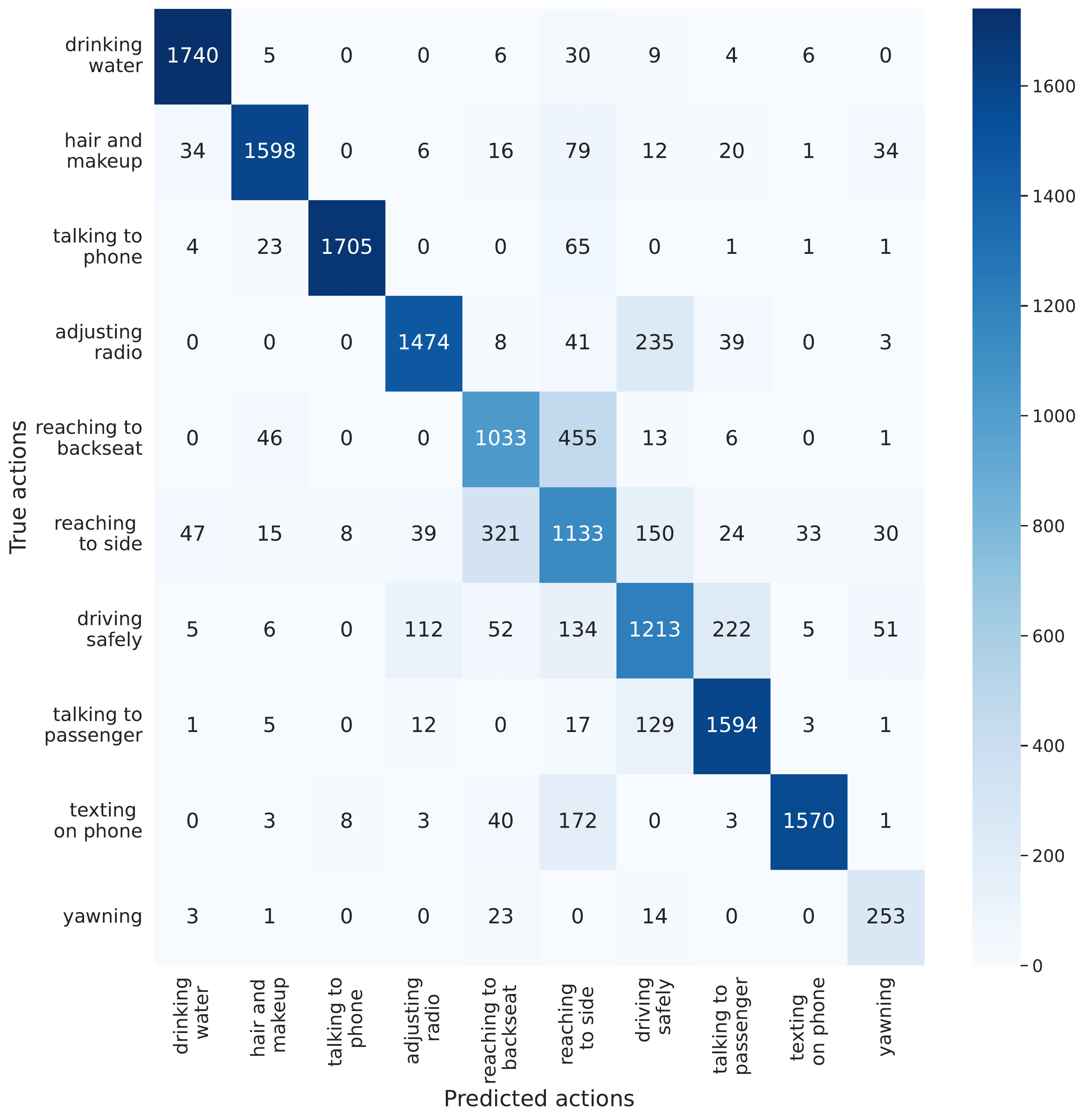}
\caption{Confusion matrix of Single-frameCLIP model on DMD dataset. It shows that the ``reaching to backseat", ``reaching to side" and ``driving safely" classes were the most challenging for the Single-frameCLIP model. Also, the dataset has limited ``yawning" samples compared to the other classes. Additionally, we noticed that only two SAM-DD classes ``head dropping" and ``touching hair" got F1-scores below 0.70 among the ten classes.}
\label{fig:clip-conf-matrix}
\end{figure}

We illustrate the class-level F1-scores of Single-frameCLIP on the DMD and SAM-DD datasets in Table~\ref{class-level} to understand the most challenging actions to classify. The average F1-scores on DMD and SAM-DD datasets are 0.819 and 0.847, respectively. In addition, we demonstrate the confusion matrix of Single-frameCLIP on fold02 of the DMD dataset as an example in Fig.~\ref{fig:clip-conf-matrix}. We have found that the DMD classes such as ``reaching to backseat", ``reaching to side" and ``driving safely" exhibited lower F1-scores ($<$0.70), indicating that these categories pose greater challenges for the framework to classify correctly. These action classes appeared to generate similar image features, leading to a higher rate of misclassification. Additionally, we noticed that only two SAM-DD classes ``head dropping" and ``touching hair" got F1-scores below 0.70 among the ten classes. To further illustrate this, Fig.~\ref{fig:missed_classified_samples} provides visual examples of instances where the Single-frameCLIP model incorrectly classified actions, comparing the model's outputs against the ground truth.

\begin{table*}[htb]
\begin{center}
\setlength{\abovecaptionskip}{2pt}
\caption{Single-frame CLIP: Removing vs. merging classes results on SynDD1 dataset}
\label{removevsmergecls}
\begin{tabular}{*{11}{c}}

\toprule Experiment & No.of  & No.of & adjusting & drinking & picking & reaching & talking & sing+eat & Mean & Mean \\
type & class & camera-view & hair & water & something & behind & phone & +yawn & F1 & accuracy\\
\midrule Removing & 5 & 3 & 0.64 & 0.71 & 0.84 & 0.82 & 0.80 & - & 0.763 & $75.62 \%$  \\

\midrule Merging & 6 & 3 & 0.51 & 0.50 & 0.82 & 0.75 & 0.88 & 0.73 & 0.698 & $69.41 \%$  \\

\midrule \begin{tabular}{@{}c@{}}Removing with \\ dashboard-view\end{tabular} & 5 & 1 & 0.77 & 0.87 & 0.89 & 0.94 & 0.89 & - & \textbf{0.873} & $\textbf{87.08\%}$  \\
\bottomrule
\end{tabular}
\end{center}
\end{table*} 

\begin{table*}[htb]
\begin{center}
\setlength{\abovecaptionskip}{2pt}
\caption{Multi-frameCLIP at different sampling rates on SynDD1 dataset (with leaving a set of driver out)}
\label{tab:Multi-frameCLIP-FPS}
\begin{tabular}{*{10}{c}}

\toprule Sampling Rate & ViT-L/14 & ViT-B/16 & RN101 & ViT-B/32 & ResNet18 & ResNet50  & Modilenetv2 & AlexNet & VGG16 \\
\midrule  0.5FPS & 56.67\% & 52.98\% & 48.57\% & 39.52\% & 43.51\% & 39.52\% & 41.31\% & 12.62\% & 12.62\% \\

\midrule 1FPS & \textbf{70.36\%} & 60.01\% & 55.60\% & 41.65\% & 47.55\% & 43.49\% & 33.65\% & 12.56\% & 12.56\%  \\

\midrule 2FPS & 59.46\% & 53.04\% & 50.36\% & 37.68\% & 40.59\% & 38.75\% & 34.11\% & 12.62\% & 12.62\%  \\

\midrule 5FPS & 62.08\% & 59.34\% & 49.34\%  & 36.96\% & 43.21\% & 39.70\% & 48.75\% & 12.62\% & 12.62\% \\
\midrule 20FPS & 61.25\% & 61.13\% & 49.35\% & 36.96\%  & 46.79\% & 40.71\% & 45.12\% & 12.62\% & 12.62\% \\
\bottomrule
\end{tabular}
\end{center}
\end{table*}

\begin{table*}[htb]
\begin{center}
\setlength{\abovecaptionskip}{2pt}
\caption{VideoCLIP at different sampling rates on SynDD1 dataset (with leaving a set of driver out)}
\label{tab:VideoCLIP-FPS}
\begin{tabular}{*{9}{c}}
\toprule
Sampling Rate   & Fold0            & Fold1            & Fold2            & Fold3            & Fold4            & Fold5            & Fold6            & Mean             \\ \midrule
1FPS  & \textbf{93.75\%} & 87.50\% & 87.50\%          & 75.00\%          & 60.00\%          & 50.00\%          & 68.75\%          & 74.64\%          \\ \midrule
2FPS  & 81.25\%          & 87.50\%          & 87.50\%          & 87.50\%          & 66.67\% & 68.75\%        & 75.00\%          & 79.17\%          \\ \midrule
5FPS  & 87.50\%          & 81.25\%          & \textbf{93.75\%} & 75.00\% & 66.67\%          & 56.25\% & 68.75\% & 75.60\% \\ \midrule
20FPS & 87.50\%          & \textbf{93.75\%}          & 87.50\%          & \textbf{93.75\%}          & \textbf{66.67\%}          & \textbf{68.75\%}          & \textbf{75.00\%}          & \textbf{81.85\%}          \\ \bottomrule
\end{tabular}
\end{center}
\end{table*}

\subsubsection{Merging and Removing Classes}
To fully understand the impact of the challenging classes on overall model performance, we pursue two additional experiments: merging and removing classes. Our initial focus was to explore how the overall model performance is influenced by either merging or removing classes that are more challenging to predict accurately. From our previous analysis, we noticed that the Single-frameCLIP model has difficulty in accurately predicting the ``singing", ``eating" and ``yawning" classes on the SynDD1 dataset compared to other classes. Table~\ref{removevsmergecls} reflects our findings, which are categorized into three different experimental settings. The first row shows the outcome following the removal of the three challenging classes with three camera views. This modification resulted in an average F1 score of 0.763 and a mean accuracy of 75.62\%. The second row illustrates the outcome of merging the three challenging classes with three camera views. This scenario led to a mean F1 score of 0.698 and an average accuracy of 69.41\%. The third row provides the results obtained from an experiment in which the three challenging classes were removed and only the dashboard camera view was considered. This experimental setup led to an average F1 score of 0.873 and a mean accuracy of 87.08\%. In conclusion, the optimization of the camera views, combined with the adjustment of the challenging classes, resulted in a significant improvement in the model's performance.

\subsubsection{Sampling Rate}
To see the effect of sampling rates on model performance, we test Multi-frameCLIP and VideoCLIP at various sampling rates on the SynDD1 dataset. Table~\ref{tab:Multi-frameCLIP-FPS} and Table~\ref{tab:VideoCLIP-FPS} show the results. We maintain the same experimental settings and observe distinct trends across different sampling rates. The ViT-L/14-based Multi-frameCLIP achieves the peak Top-1 accuracy at a 1 FPS sampling rate with an accuracy of 70.36\%. This was the highest accuracy achieved across all tested models and sampling rate settings for this experiment. In comparison, the AlexNet and VGG16 models got the lowest accuracy values. However, in Table~\ref{tab:VideoCLIP-FPS} the VideoCLIP achieves the highest Top-1 accuracy at 20 FPS, with an accuracy of 81.85\%. This contrast highlights the models' sensitivity to sampling rate variations, suggesting that the optimal rate depends on the specific model and dataset characteristics. The changes from one frame to the next are more significant at a 1 FPS sampling rate, which helped the Multi-frameCLIP model make more accurate predictions. 
Therefore, 1 FPS is the optimal sampling rate for the Multi-frameCLIP model under the conditions of our study.
On the other hand, as the average action duration for the SynDD1 dataset is $\sim$18.7 sec, a 20 FPS rate results in a better temporal resolution, allowing the VideoCLIP model to capture more subtle changes and movements. If the sampling rate is too small, then the video clip will span more than the average action duration and confuse the model prediction. Therefore, the impact of the sampling rate would depend on the average action duration and the characteristics of the downstream tasks.

\subsubsection{Effect of Camera Views}
\begin{table}[htb]
\begin{center}
\setlength{\abovecaptionskip}{2pt}
\caption{Effect of Camera-view on SAM-DD dataset}
\label{camera-view-result}
\begin{tabular}{*{5}{c}}
\toprule
Models &Camera& Top-1 & Leaving a set\\
(ours) &view & accuracy &of driver out\\
\midrule Single-frameCLIP &Dashboard&88.919&\checkmark   \\
&Side-view&\textbf{89.142}&\checkmark\\
\midrule Multi-frameCLIP &Dashboard&88.908&\checkmark   \\
&Side-view&\textbf{89.142}&\checkmark\\
\bottomrule
\end{tabular}
\end{center}
\end{table}
In this section, we investigate the effect of camera position on model performance. Table~\ref{camera-view-result} presents the Top-1 accuracy for Single-frameCLIP and Multi-frameCLIP on the SAM-DD dataset, considering two distinct camera views. The outcomes suggest that the side-view yields superior results when compared to the dashboard camera view. These results underscore the significance of the camera's perspective in detecting driver distractions. The side camera view appears to provide the most effective angle for the SAM-DD dataset, successfully capturing the driver's movements and body maneuvers.

\subsection{Computational Complexity}
\begin{table}[htbp]
\setlength{\abovecaptionskip}{-2pt}
\begin{center}
\setlength{\abovecaptionskip}{-5pt}
\caption{VideoCLIP computation efficiency (Batch size=1)}
\label{tab:VideoCLIP-FLOPS}
\resizebox{\columnwidth}{!}{%
\begin{tabular}{*{6}{c}}
\toprule
FLOPs &\#Params.& Forw./backw. & Params& Estimated\\
(G) &(M) & pass size(MB) &size (MB)& Total size(MB)\\
\midrule 3.54 &9.67&964.26&38.68&1021  \\
\bottomrule
\end{tabular}}
\end{center}
\end{table}

In Table~\ref{tab:VideoCLIP-FLOPS}, we present the details of the computational efficacy of the proposed VideoCLIP model. In terms of driver distraction recognition tasks, we compared the floating point operations (FLOPs) and the number of parameters ($\#$Params.) of VideoCLIP with its counterparts. VideoCLIP demonstrates superior FLOPs efficiency and a relatively smaller parameter count ($<$10M). For example, in the SAM-DD dataset~\cite{sam-dd-dataset} setting, Swin-T gets 4.5G FLOPs and 28.3M trainable parameters. In~\cite{aicity23top} setting, VideoMAE operates at 57G FLOPs with 22M parameters, ResNet50~\cite{rn50} uses 3.8 GFLOPs with 25.6M parameters for an input size of 224x224x3. Therefore, the computational performance of the VideoCLIP model coupled with a Top-1 accuracy of 81.85\% makes it a resource-effective model in the driver distraction recognition task context. This efficiency is particularly valuable in real-time Intelligent Transportation applications like driver distraction recognition, where quick and accurate video processing is essential for safety and effectiveness. Moreover, the small parameter count also implies potentially lower memory requirements and faster training times, further enhancing its applicability in practical scenarios.

\subsection{Anecdotal Observation}
In this section, we discuss the cases where our proposed frameworks struggle to perform well and investigate the possible reasons.

\subsubsection{Failure cases}
\label{sec:Failure_cases}
\begin{figure}[htb]
\centering
\includegraphics[width=3.2in]{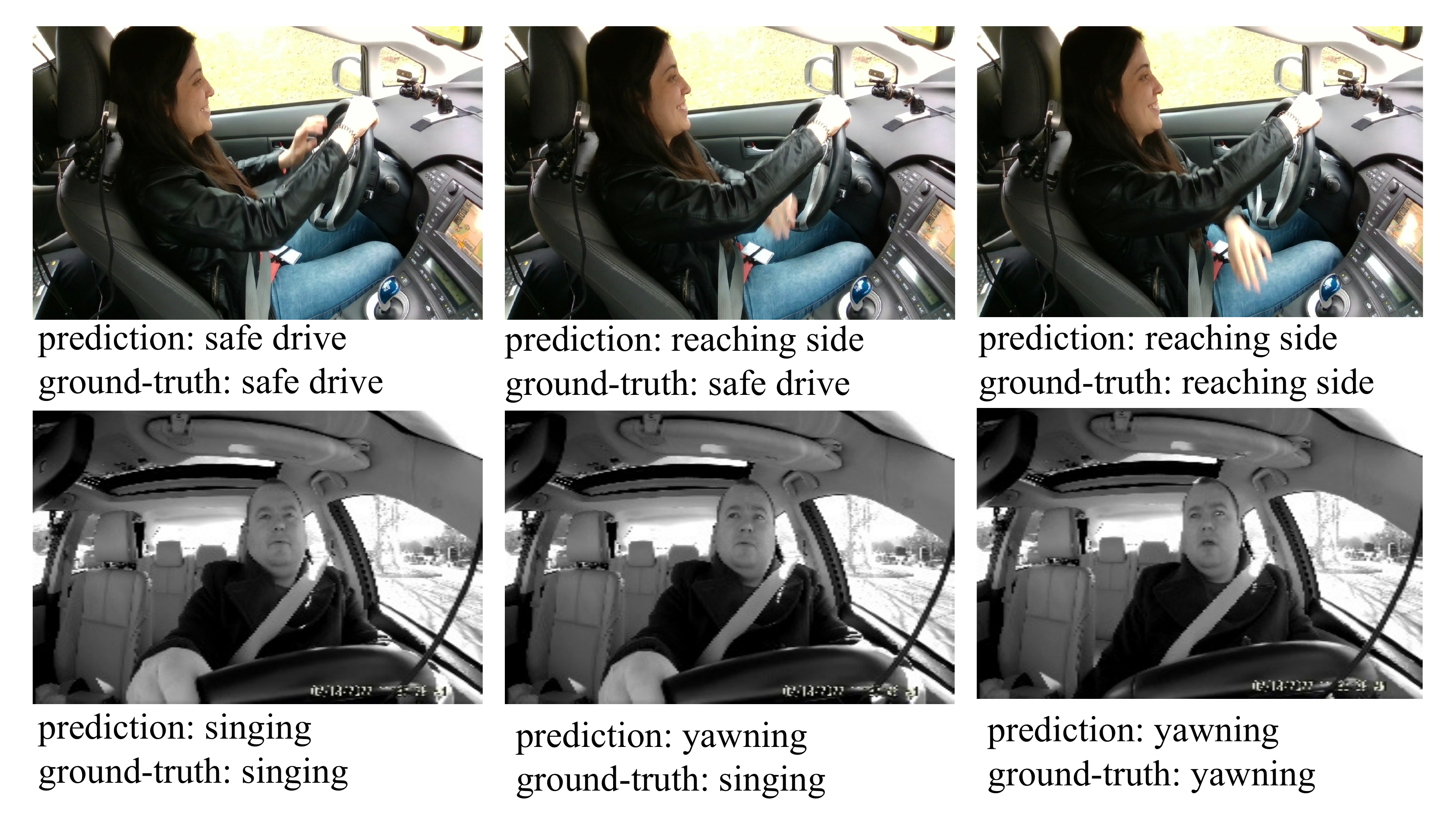}
\caption{Some example frames that were misclassified by the Single-frameCLIP on DMD (top row) and SynDD1 (bottom row). These examples demonstrate how Single-frameCLIP fails to capture the temporal information only by looking at a single frame at a time.}
\label{fig:missed_classified_samples}
\end{figure}

We demonstrate some miss-classified frames from the Single-frameCLIP model on the DMD and SynDD1 dataset in Fig.~\ref{fig:missed_classified_samples}. Analyzing the confusion matrix and the miss-classified frames, it is evident that the most challenging classes for the model occur where there is considerable feature overlap between classes that exhibit lower F1-scores. The middle column of Fig~\ref{fig:missed_classified_samples}
shows the overlapping cues between two actions characterized by specific hand movements or facial expressions.

For example, the first-row middle frame driver's left-hand position close to the steering wheel resembles the starting motion to the side, which confused the model to incorrectly predict the ``reaching side" class. Therefore, the Single-frameCLIP model struggles to distinguish actions where the visual movements are subtle and there is a slight temporal context. Most of the time, it is quite complicated to decipher these activities only by looking at a single frame. To address this issue, the Multi-frameCLIP and VideoCLIP frameworks were developed, where a sequence of consecutive frames is assessed for a more accurate prediction score. Moreover, these models leverage the temporal information across multiple frames and capture subtle changes in hand movements and facial gestures. This consideration contributed to a significant performance enhancement relative to the Single-frameCLIP model.


\subsubsection{Frame-by-frame Analysis on SynDD1 Video Clips}
\begin{figure*}[t]
\centering
\includegraphics[width=5in]{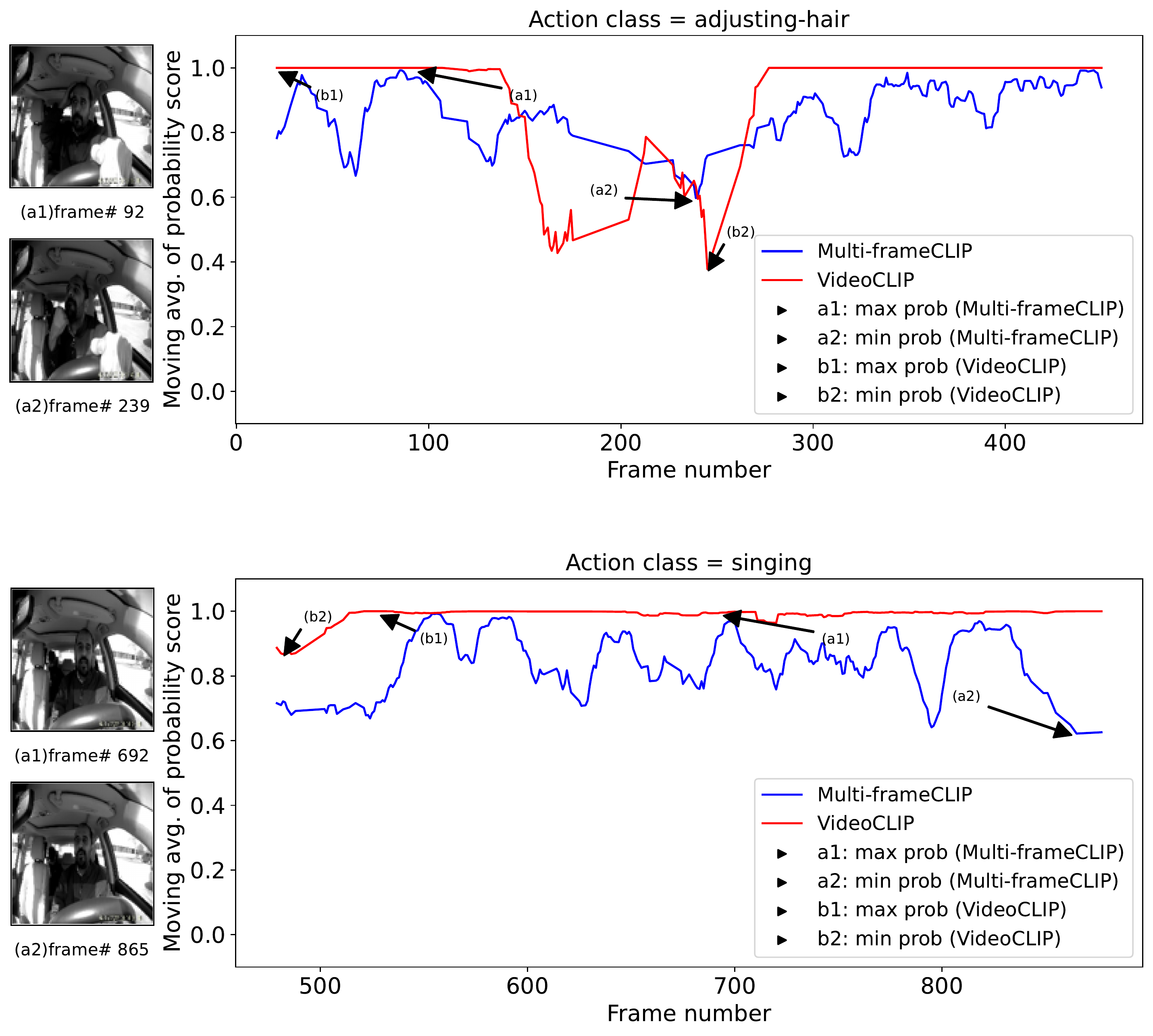}
\caption{This plot shows the predicted probability score vs. frame number. We analyze the Multi-frameCLIP and VideoCLIP model's prediction probability scores for the entire action. We notice distinct changes in frames for high and low probability scores. This figure demonstrates that VideoCLIP outperforms Multi-frameCLIP's prediction scores, offering a better approach to capturing temporal features. Therefore, VideoCLIP performs better in classifying actions on video data }
\label{frame-by-frame-pred-for-video}
\end{figure*}

Fig.~\ref{frame-by-frame-pred-for-video} presents the frame-by-frame analysis of driverID 25470 from the SynDD1 dataset using the Multi-frameCLIP and VideoCLIP. It demonstrates the dynamic probabilities of predictions produced by the Multi-frameCLIP and VideoCLIP models over the entire span of an action. The vertical axis shows the moving average of the computed probabilities against the individual frame numbers, represented by the horizontal axis. We observe distinct fluctuations between high and low probability scores, aligning with frames where the model either successfully or unsuccessfully classified the given action. Our analysis focuses on two distinct actions - adjusting hair and singing, at a sampling rate of 20 FPS. The representation of lower and higher probability scores from both models is clearly depicted in this plot. An important observation is that the VideoCLIP demonstrates a stable and consistent prediction score throughout the entire video compared to that of the Multi-frameCLIP. Therefore, we conclude that VideoCLIP has the capability to efficiently capture temporal information from videos, significantly improving prediction accuracy while maintaining consistency. Moreover, the plot illustrates the superiority of both Multi-frameCLIP and VideoCLIP in extracting the temporal correlations among frames, thus providing insights into continuous actions that are challenging to get from isolated frames, as in the Single-frameCLIP scenario.


\subsection{Limitaions and Future directions}
Our algorithm shows promising results within its current setup, yet there is potential to further its adaptability and performance in varied real-world driving conditions across different unusual distraction categories. The variety of distractions that drivers encounter is vast and often unpredictable. Therefore, we need a sophisticated monitoring system to recognize the natural interactions between humans and vehicles and encourage safe driving practices. This system must be equipped to accurately identify the driver's current state and assess the reliability of its predictions. Our current models were tested under controlled and available distracted settings from public datasets. The performance and efficacy in real-world applications, considering diverse hardware and environmental conditions, were not fully explored. To enhance the robustness and applicability of our algorithm, future work could involve training with additional data that cover a wider range of driving conditions -  multiple distracted behaviors occurring simultaneously or alternately, unexpected or out-of-distribution behavior categories, occlusion events and so on. This would help in validating the algorithm's effectiveness across various scenarios. Optimizing the algorithm for efficient performance across a range of hardware platforms, including low-power devices or with limited compute resources, could be a significant area of focus. This would make the system more viable for real-time applications in vehicles. Additionally, future research should also consider the ethical implications and privacy concerns associated with driver monitoring, ensuring compliance with relevant regulations and societal norms. Also, we plan to address the misclassification issues and enhance the reliability of the model predictions by implementing reliable uncertainty estimates~\cite{un-aware}. This would be particularly beneficial in complex or ambiguous scenarios where driver behavior is not distinctly classifiable, thus allowing for a more nuanced understanding and interpretation of the model's confidence measures. We believe that this addition will enhance the safety and reliability of the automated driver assistance systems by providing a more comprehensive and trustworthy analysis of driver behavior.

\section{Conclusion}
\label{sec:conclusion}


In this paper, we introduce a vision-language framework for recognizing distracted driving behaviors from naturalistic driving videos.
Our findings reveal that task-based fine-tuning enhances the Single-frameCLIP model's accuracy, while Multi-frameCLIP and VideoCLIP models can leverage the temporal contexts of a video and significantly outperform the frame-based models.
Our proposed frameworks exhibit state-of-the-art performance at distracted action recognition and outperform the recent benchmark models.  
This study emphasizes the real-world applicability and generalizability of these models while also pointing out areas for improvement. Future work will focus on optimizing visual-textual relationships, integrating multiple camera views for comprehensive behavior analysis, and testing the model's generalizability across various real-world scenarios. 
Moreover, the generalizability of our model across various scenarios and datasets presents an intriguing prospect. 
In addition, we recognize the importance of empirical validation and testing our system under more challenging, unexpected, and out-of-distribution driving behavior categories. To this end, we are actively planning an extension of our current work and aiming to extend the model's testing to include a broader spectrum of actions, particularly focusing on those that are rare or challenging.
These tests are intended to validate our framework's robustness and enhance its adaptive capabilities further. Finally, systematic evaluations under diverse environmental conditions, such as varying lighting conditions and data quality, could offer valuable insights into the model's robustness and ability to function optimally under real-world driving circumstances.


\bibliographystyle{IEEEtran}
\bibliography{main}

\begin{thebibliography}{10}
\providecommand{\url}[1]{#1}
\csname url@samestyle\endcsname
\providecommand{\newblock}{\relax}
\providecommand{\bibinfo}[2]{#2}
\providecommand{\BIBentrySTDinterwordspacing}{\spaceskip=0pt\relax}
\providecommand{\BIBentryALTinterwordstretchfactor}{4}
\providecommand{\BIBentryALTinterwordspacing}{\spaceskip=\fontdimen2\font plus
\BIBentryALTinterwordstretchfactor\fontdimen3\font minus \fontdimen4\font\relax}
\providecommand{\BIBforeignlanguage}[2]{{%
\expandafter\ifx\csname l@#1\endcsname\relax
\typeout{** WARNING: IEEEtran.bst: No hyphenation pattern has been}%
\typeout{** loaded for the language `#1'. Using the pattern for}%
\typeout{** the default language instead.}%
\else
\language=\csname l@#1\endcsname
\fi
#2}}
\providecommand{\BIBdecl}{\relax}
\BIBdecl

\bibitem{NHTSA2023}
{National Center for Statistics and Analysis}, ``\textit{Distracted Driving in 2021},'' National Highway Traffic Safety Administration, Research Note DOT HS 813 443, May 2023.

\bibitem{WHO2023}
``Global status report on road safety 2023,'' World Health Organization, Geneva, Safety and Mobility (SAM), Social Determinants of Health (SDH)~81, 2023, licence: CC BY-NC-SA 3.0 IGO.

\bibitem{cdc}
{National Highway Traffic Safety Administration}, ``\textit{Overview of the National Highway Traffic Safety Administration's Driver Distraction Program},'' 2010, {U.S.} Department of Transportation, Washington, {DC.} 8 February 2022.

\bibitem{dingus2016driver}
T.~A. Dingus, F.~Guo, S.~Lee, J.~F. Antin, M.~Perez, M.~Buchanan-King, and J.~Hankey, ``Driver crash risk factors and prevalence evaluation using naturalistic driving data,'' \emph{Proceedings of the National Academy of Sciences}, vol. 113, no.~10, pp. 2636--2641, 2016.

\bibitem{klauer2006}
S.~G. Klauer, T.~A. Dingus, V.~L. Neale, J.~D. Sudweeks, and D.~J. Ramsey, ``The impact of driver inattention on near-crash/crash risk: An analysis using the 100-car naturalistic driving study data.'' United States. National Highway Traffic Safety Administration, Washington, DC, FHWA-HRT-04-138, 2006.

\bibitem{arvin2019}
R.~Arvin, M.~Kamrani, and A.~J. Khattak, ``The role of pre-crash driving instability in contributing to crash intensity using naturalistic driving data,'' \emph{Accident Analysis \& Prevention}, vol. 132, p. 105226, 2019.

\bibitem{arvin2020}
R.~Arvin and A.~J. Khattak, ``Driving impairments and duration of distractions: assessing crash risk by harnessing microscopic naturalistic driving data,'' \emph{Accident Analysis \& Prevention}, vol. 146, p. 105733, 2020.

\bibitem{young2007}
K.~Young and M.~Regan, ``Driver distraction: A review of the literature,'' in \emph{Distracted Driving}, I.~Faulks, M.~Regan, M.~Stevenson, J.~Brown, A.~Porter, and J.~Irwin, Eds.\hskip 1em plus 0.5em minus 0.4em\relax Sydney, NSW: Australasian College of Road Safety, 2007, pp. 379--405.

\bibitem{kashevnik2021}
A.~Kashevnik, R.~Shchedrin, C.~Kaiser, and A.~Stocker, ``Driver distraction detection methods: A literature review and framework,'' \emph{IEEE Access}, vol.~9, pp. 60\,063--60\,076, 2021.

\bibitem{Survey}
J.~Wang, W.~Chai, A.~Venkatachalapathy, K.~L. Tan, A.~Haghighat, S.~Velipasalar, Y.~Adu-Gyamfi, and A.~Sharma, ``A survey on driver behavior analysis from in-vehicle cameras,'' \emph{IEEE Transactions on Intelligent Transportation Systems}, vol.~23, no.~8, pp. 10\,186--10\,209, 2022.

\bibitem{clip}
A.~Radford, J.~W. Kim, C.~Hallacy, A.~Ramesh, G.~Goh, S.~Agarwal, G.~Sastry, A.~Askell, P.~Mishkin, J.~Clark \emph{et~al.}, ``Learning transferable visual models from natural language supervision,'' in \emph{International conference on machine learning}.\hskip 1em plus 0.5em minus 0.4em\relax PMLR, 2021, pp. 8748--8763.

\bibitem{dmd-dataset}
J.~D. Ortega, N.~Kose, P.~Cañas, M.-A. Chao, A.~Unnervik, M.~Nieto, O.~Otaegui, and L.~Salgado, ``Dmd: A large-scale multi-modal driver monitoring dataset for attention and alertness analysis,'' in \emph{Computer Vision -- ECCV 2020 Workshops}, A.~Bartoli and A.~Fusiello, Eds.\hskip 1em plus 0.5em minus 0.4em\relax Springer International Publishing, 2020, pp. 387--405.

\bibitem{dad}
O.~Kopuklu, J.~Zheng, H.~Xu, and G.~Rigoll, ``Driver anomaly detection: A dataset and contrastive learning approach,'' in \emph{Proceedings of the IEEE/CVF Winter Conference on Applications of Computer Vision}, 2021, pp. 91--100.

\bibitem{DHCNN}
B.~Qin, J.~Qian, Y.~Xin, B.~Liu, and Y.~Dong, ``Distracted driver detection based on a cnn with decreasing filter size,'' \emph{IEEE Transactions on Intelligent Transportation Systems}, vol.~23, no.~7, pp. 6922--6933, 2022.

\bibitem{Faiqa2021}
F.~Sajid, A.~R. Javed, A.~Basharat, N.~Kryvinska, A.~Afzal, and M.~Rizwan, ``An efficient deep learning framework for distracted driver detection,'' \emph{IEEE Access}, vol.~9, pp. 169\,270--169\,280, 2021.

\bibitem{3dcnn2019}
N.~Moslemi, R.~Azmi, and M.~Soryani, ``Driver {D}istraction {R}ecognition using {3D} {C}onvolutional {N}eural {N}etworks,'' in \emph{2019 4th International Conference on Pattern Recognition and Image Analysis (IPRIA)}, 2019, pp. 145--151.

\bibitem{unsupervised-Li}
B.~Li, J.~Chen, Z.~Huang, H.~Wang, J.~Lv, J.~Xi, J.~Zhang, and Z.~Wu, ``A new unsupervised deep learning algorithm for fine-grained detection of driver distraction,'' \emph{IEEE Transactions on Intelligent Transportation Systems}, vol.~23, no.~10, pp. 19\,272--19\,284, 2022.

\bibitem{sam-dd-dataset}
H.~Yang, H.~Liu, Z.~Hu, A.-T. Nguyen, T.-M. Guerra, and C.~Lv, ``Quantitative identification of driver distraction: A weakly supervised contrastive learning approach,'' \emph{IEEE Transactions on Intelligent Transportation Systems}, pp. 1--12, 2023.

\bibitem{videoclip}
H.~Xu, G.~Ghosh, P.-Y. Huang, D.~Okhonko, A.~Aghajanyan, F.~Metze, L.~Zettlemoyer, and C.~Feichtenhofer, ``{VideoCLIP}: Contrastive pre-training for zero-shot video-text understanding,'' in \emph{Proceedings of the 2021 Conference on Empirical Methods in Natural Language Processing (EMNLP)}.\hskip 1em plus 0.5em minus 0.4em\relax Association for Computational Linguistics, Nov. 2021.

\bibitem{actionclip}
M.~Wang, J.~Xing, and Y.~Liu, ``Actionclip: A new paradigm for video action recognition,'' \emph{arXiv preprint arXiv:2109.08472}, 2021.

\bibitem{dalle}
A.~Ramesh, M.~Pavlov, G.~Goh, S.~Gray, C.~Voss, A.~Radford, M.~Chen, and I.~Sutskever, ``Zero-shot text-to-image generation,'' in \emph{Proceedings of the 38th International Conference on Machine Learning}, ser. Proceedings of Machine Learning Research, M.~Meila and T.~Zhang, Eds., vol. 139.\hskip 1em plus 0.5em minus 0.4em\relax PMLR, 18--24 Jul 2021, pp. 8821--8831.

\bibitem{flava}
A.~Singh, R.~Hu, V.~Goswami, G.~Couairon, W.~Galuba, M.~Rohrbach, and D.~Kiela, ``Flava: A foundational language and vision alignment model,'' in \emph{Proceedings of the IEEE/CVF Conference on Computer Vision and Pattern Recognition}, 2022, pp. 15\,638--15\,650.

\bibitem{VLMo}
H.~Bao, W.~Wang, L.~Dong, Q.~Liu, O.~K. Mohammed, K.~Aggarwal, S.~Som, S.~Piao, and F.~Wei, ``{VLM}o: Unified vision-language pre-training with mixture-of-modality-experts,'' in \emph{Advances in Neural Information Processing Systems}, A.~H. Oh, A.~Agarwal, D.~Belgrave, and K.~Cho, Eds., 2022.

\bibitem{VL-t5}
J.~Cho, J.~Lei, H.~Tan, and M.~Bansal, ``Unifying vision-and-language tasks via text generation,'' in \emph{Proceedings of the 38th International Conference on Machine Learning}, ser. Proceedings of Machine Learning Research, M.~Meila and T.~Zhang, Eds., vol. 139.\hskip 1em plus 0.5em minus 0.4em\relax PMLR, 18--24 Jul 2021, pp. 1931--1942.

\bibitem{detic}
X.~Zhou, R.~Girdhar, A.~Joulin, P.~Kr{\"a}henb{\"u}hl, and I.~Misra, ``Detecting twenty-thousand classes using image-level supervision,'' in \emph{Computer Vision--ECCV 2022: 17th European Conference, Tel Aviv, Israel, October 23--27, 2022, Proceedings, Part IX}.\hskip 1em plus 0.5em minus 0.4em\relax Springer, 2022, pp. 350--368.

\bibitem{feuer2023zero}
B.~Feuer, A.~Joshi, M.~Cho, K.~Jani, S.~Chiranjeevi, Z.~K. Deng, A.~Balu, A.~K. Singh, S.~Sarkar, N.~Merchant, A.~Singh, B.~Ganapathysubramanian, and C.~Hegde, ``Zero-shot insect detection via weak language supervision,'' in \emph{2nd AAAI Workshop on AI for Agriculture and Food Systems}, 2023.

\bibitem{imgCLIP-to-vid}
H.~Fang, P.~Xiong, L.~Xu, and W.~Luo, ``Transferring image-clip to video-text retrieval via temporal relations,'' \emph{IEEE Transactions on Multimedia}, vol.~25, pp. 7772--7785, 2023.

\bibitem{bottle-neck-video-text}
M.~Patrick, P.-Y. Huang, Y.~Asano, F.~Metze, A.~G. Hauptmann, J.~F. Henriques, and A.~Vedaldi, ``Support-set bottlenecks for video-text representation learning,'' in \emph{International Conference on Learning Representations}, 2021.

\bibitem{coot}
S.~Ging, M.~Zolfaghari, H.~Pirsiavash, and T.~Brox, ``Coot: Cooperative hierarchical transformer for video-text representation learning,'' in \emph{Advances in Neural Information Processing Systems}, H.~Larochelle, M.~Ranzato, R.~Hadsell, M.~Balcan, and H.~Lin, Eds., vol.~33.\hskip 1em plus 0.5em minus 0.4em\relax Curran Associates, Inc., 2020, pp. 22\,605--22\,618.

\bibitem{clipforge}
A.~Sanghi, H.~Chu, J.~G. Lambourne, Y.~Wang, C.~Cheng, M.~Fumero, and K.~Malekshan, ``Clip-forge: Towards zero-shot text-to-shape generation,'' in \emph{2022 IEEE/CVF Conference on Computer Vision and Pattern Recognition (CVPR)}.\hskip 1em plus 0.5em minus 0.4em\relax Los Alamitos, CA, USA: IEEE Computer Society, jun 2022, pp. 18\,582--18\,592.

\bibitem{jain2021dreamfields}
A.~Jain, B.~Mildenhall, J.~T. Barron, P.~Abbeel, and B.~Poole, ``Zero-shot text-guided object generation with dream fields,'' \emph{CVPR}, 2022.

\bibitem{clip2vid}
H.~Fang, P.~Xiong, L.~Xu, and Y.~Chen, ``Clip2video: Mastering video-text retrieval via image clip,'' \emph{arXiv preprint arXiv:2106.11097}, 2021.

\bibitem{weiheng}
W.~Chai, J.~Wang, J.~Chen, S.~Velipasalar, and A.~Sharma, ``Rethinking the evaluation of driver behavior analysis approaches,'' \emph{IEEE Transactions on Intelligent Transportation Systems}, pp. 1--9, 2024.

\bibitem{body2}
M.~Martin, J.~Popp, M.~Anneken, M.~Voit, and R.~Stiefelhagen, ``Body pose and context information for driver secondary task detection,'' in \emph{2018 IEEE Intelligent Vehicles Symposium (IV)}, 2018, pp. 2015--2021.

\bibitem{hand1}
N.~Das, E.~Ohn-Bar, and M.~M. Trivedi, ``On performance evaluation of driver hand detection algorithms: Challenges, dataset, and metrics,'' in \emph{2015 IEEE 18th International Conference on Intelligent Transportation Systems}, 2015, pp. 2953--2958.

\bibitem{eye-gaze}
L.~Yang, K.~Dong, A.~J. Dmitruk, J.~Brighton, and Y.~Zhao, ``A dual-cameras-based driver gaze mapping system with an application on non-driving activities monitoring,'' \emph{IEEE Transactions on Intelligent Transportation Systems}, vol.~21, no.~10, pp. 4318--4327, 2020.

\bibitem{drozy-data}
Q.~Massoz, T.~Langohr, C.~François, and J.~G. Verly, ``The ulg multimodality drowsiness database (called drozy) and examples of use,'' in \emph{2016 IEEE Winter Conference on Applications of Computer Vision (WACV)}, 2016, pp. 1--7.

\bibitem{eeg-LSTM}
X.~Zuo, C.~Zhang, F.~Cong, J.~Zhao, and T.~Hämäläinen, ``Driver distraction detection using bidirectional long short-term network based on multiscale entropy of eeg,'' \emph{IEEE Transactions on Intelligent Transportation Systems}, vol.~23, no.~10, pp. 19\,309--19\,322, 2022.

\bibitem{ml-feature}
E.~Ohn-Bar and M.~M. Trivedi, ``Looking at humans in the age of self-driving and highly automated vehicles,'' \emph{IEEE Transactions on Intelligent Vehicles}, vol.~1, no.~1, pp. 90--104, 2016.

\bibitem{HMM}
A.~Jain, H.~S. Koppula, B.~Raghavan, S.~Soh, and A.~Saxena, ``Car that knows before you do: Anticipating maneuvers via learning temporal driving models,'' in \emph{2015 IEEE International Conference on Computer Vision (ICCV)}.\hskip 1em plus 0.5em minus 0.4em\relax Los Alamitos, CA, USA: IEEE Computer Society, dec 2015, pp. 3182--3190.

\bibitem{random-forest}
K.~Ben~Ahmed, B.~Goel, P.~Bharti, S.~Chellappan, and M.~Bouhorma, ``Leveraging smartphone sensors to detect distracted driving activities,'' \emph{IEEE Transactions on Intelligent Transportation Systems}, vol.~20, no.~9, pp. 3303--3312, 2019.

\bibitem{knn}
R.~Khosrowabadi, M.~Heijnen, A.~Wahab, and H.~C. Quek, ``The dynamic emotion recognition system based on functional connectivity of brain regions,'' in \emph{2010 IEEE Intelligent Vehicles Symposium}, 2010, pp. 377--381.

\bibitem{toward}
D.~Liu, T.~Yamasaki, Y.~Wang, K.~Mase, and J.~Kato, ``Toward extremely lightweight distracted driver recognition with distillation-based neural architecture search and knowledge transfer,'' \emph{IEEE Transactions on Intelligent Transportation Systems}, vol.~24, no.~1, pp. 764--777, 2023.

\bibitem{FRNet}
C.~Duan, Y.~Gong, J.~Liao, M.~Zhang, and L.~Cao, ``Frnet: Dcnn for real-time distracted driving detection toward embedded deployment,'' \emph{IEEE Transactions on Intelligent Transportation Systems}, vol.~24, no.~9, pp. 9835--9848, 2023.

\bibitem{facetracking}
X.~Shi, S.~Shan, M.~Kan, S.~Wu, and X.~Chen, ``Real-time rotation-invariant face detection with progressive calibration networks,'' in \emph{Proceedings of the IEEE Conference on Computer Vision and Pattern Recognition}, 2018, pp. 2295--2303.

\bibitem{opticalflow}
N.~Kose, O.~Kopuklu, A.~Unnervik, and G.~Rigoll, ``Real-time driver state monitoring using a cnn based spatio-temporal approach,'' in \emph{2019 IEEE Intelligent Transportation Systems Conference (ITSC)}.\hskip 1em plus 0.5em minus 0.4em\relax IEEE, 2019, pp. 3236--3242.

\bibitem{skpose}
P.~Li, M.~Lu, Z.~Zhang, D.~Shan, and Y.~Yang, ``A novel spatial-temporal graph for skeleton-based driver action recognition,'' in \emph{2019 IEEE Intelligent Transportation Systems Conference (ITSC)}.\hskip 1em plus 0.5em minus 0.4em\relax IEEE, 2019, pp. 3243--3248.

\bibitem{vgg16}
K.~Simonyan and A.~Zisserman, ``Very deep convolutional networks for large-scale image recognition,'' in \emph{International Conference on Learning Representations}, 2015.

\bibitem{rn50}
K.~He, X.~Zhang, S.~Ren, and J.~Sun, ``Deep residual learning for image recognition,'' in \emph{2016 IEEE Conference on Computer Vision and Pattern Recognition (CVPR)}, 2016, pp. 770--778.

\bibitem{AlexNet}
A.~Krizhevsky, I.~Sutskever, and G.~E. Hinton, ``Imagenet classification with deep convolutional neural networks,'' in \emph{Advances in Neural Information Processing Systems}, F.~Pereira, C.~Burges, L.~Bottou, and K.~Weinberger, Eds., vol.~25.\hskip 1em plus 0.5em minus 0.4em\relax Curran Associates, Inc., 2012.

\bibitem{incepv3}
C.~Szegedy, V.~Vanhoucke, S.~Ioffe, J.~Shlens, and Z.~Wojna, ``Rethinking the inception architecture for computer vision,'' in \emph{2016 IEEE Conference on Computer Vision and Pattern Recognition (CVPR)}, 2016, pp. 2818--2826.

\bibitem{mobilenetv2}
M.~Sandler, A.~Howard, M.~Zhu, A.~Zhmoginov, and L.-C. Chen, ``Mobilenetv2: Inverted residuals and linear bottlenecks,'' in \emph{Proceedings of the IEEE Conference on Computer Vision and Pattern Recognition (CVPR)}, 2018, pp. 4510--4520.

\bibitem{shufflenet}
N.~Ma, X.~Zhang, H.-T. Zheng, and J.~Sun, ``Shufflenet v2: Practical guidelines for efficient cnn architecture design,'' in \emph{Proceedings of the European conference on computer vision (ECCV)}, 2018, pp. 116--131.

\bibitem{mobilevgg}
B.~Baheti, S.~Talbar, and S.~Gajre, ``Towards computationally efficient and realtime distracted driver detection with mobilevgg network,'' \emph{IEEE Transactions on Intelligent Vehicles}, vol.~5, no.~4, pp. 565--574, 2020.

\bibitem{driveandact}
M.~Martin, A.~Roitberg, M.~Haurilet, M.~Horne, S.~Reiß, M.~Voit, and R.~Stiefelhagen, ``Driveact: A multi-modal dataset for fine-grained driver behavior recognition in autonomous vehicles,'' in \emph{2019 IEEE/CVF International Conference on Computer Vision (ICCV)}, 2019, pp. 2801--2810.

\bibitem{quo}
J.~Carreira and A.~Zisserman, ``Quo vadis, action recognition? a new model and the kinetics dataset,'' in \emph{proceedings of the IEEE Conference on Computer Vision and Pattern Recognition (CVPR)}, 2017, pp. 6299--6308.

\bibitem{3dcnndiba}
A.~Diba, M.~Fayyaz, V.~Sharma, M.~M. Arzani, R.~Yousefzadeh, J.~Gall, and L.~Van~Gool, ``Spatio-temporal channel correlation networks for action classification,'' in \emph{Proceedings of the European Conference on Computer Vision (ECCV)}, 2018, pp. 284--299.

\bibitem{D3d}
J.~Stroud, D.~Ross, C.~Sun, J.~Deng, and R.~Sukthankar, ``D3d: Distilled 3d networks for video action recognition,'' in \emph{Proceedings of the IEEE/CVF Winter Conference on Applications of Computer Vision}, 2020, pp. 625--634.

\bibitem{auc}
Y.~Abouelnaga, H.~M. Eraqi, and M.~N. Moustafa, ``Real-time distracted driver posture classification,'' \emph{arXiv preprint arXiv:1706.09498}, 2017.

\bibitem{hcf}
C.~Huang, X.~Wang, J.~Cao, S.~Wang, and Y.~Zhang, ``Hcf: a hybrid cnn framework for behavior detection of distracted drivers,'' \emph{IEEE access}, vol.~8, pp. 109\,335--109\,349, 2020.

\bibitem{ensemble}
H.~M. Eraqi, Y.~Abouelnaga, M.~H. Saad, M.~N. Moustafa \emph{et~al.}, ``Driver distraction identification with an ensemble of convolutional neural networks,'' \emph{Journal of Advanced Transportation}, vol. 2019, 2019.

\bibitem{twostream1}
K.~Simonyan and A.~Zisserman, ``Two-stream convolutional networks for action recognition in videos,'' \emph{Advances in Neural Information Processing Systems}, vol.~27, 2014.

\bibitem{twostream2}
C.~Feichtenhofer, A.~Pinz, and A.~Zisserman, ``Convolutional two-stream network fusion for video action recognition,'' in \emph{Proceedings of the IEEE Conference on Computer Vision and Pattern Recognition}, 2016, pp. 1933--1941.

\bibitem{VIT}
\BIBentryALTinterwordspacing
A.~Dosovitskiy, L.~Beyer, A.~Kolesnikov, D.~Weissenborn, X.~Zhai, T.~Unterthiner, M.~Dehghani, M.~Minderer, G.~Heigold, S.~Gelly, J.~Uszkoreit, and N.~Houlsby, ``An image is worth 16x16 words: Transformers for image recognition at scale,'' in \emph{International Conference on Learning Representations}, 2021. [Online]. Available: \url{https://openreview.net/forum?id=YicbFdNTTy}
\BIBentrySTDinterwordspacing

\bibitem{vivit}
A.~Arnab, M.~Dehghani, G.~Heigold, C.~Sun, M.~Lucic, and C.~Schmid, ``Vivit: A video vision transformer,'' in \emph{2021 IEEE/CVF International Conference on Computer Vision (ICCV)}.\hskip 1em plus 0.5em minus 0.4em\relax Los Alamos, CA, USA: IEEE Computer Society, oct 2021, pp. 6816--6826.

\bibitem{mvit}
H.~Fan, B.~Xiong, K.~Mangalam, Y.~Li, Z.~Yan, J.~Malik, and C.~Feichtenhofer, ``Multiscale vision transformers,'' in \emph{Proceedings of the IEEE/CVF International Conference on Computer Vision}, 2021, pp. 6824--6835.

\bibitem{vidt}
G.~Bertasius, H.~Wang, and L.~Torresani, ``Is space-time attention all you need for video understanding?'' in \emph{Proceedings of the 38th International Conference on Machine Learning}, vol.~2, no.~3, 2021, p.~4.

\bibitem{swin-T}
Z.~Liu, J.~Ning, Y.~Cao, Y.~Wei, Z.~Zhang, S.~Lin, and H.~Hu, ``Video swin transformer,'' in \emph{Proceedings of the IEEE/CVF Conference on Computer Vision and Pattern Recognition(CVPR)}, 2022, pp. 3202--3211.

\bibitem{dhakate2020}
K.~R. Dhakate and R.~Dash, ``Distracted driver detection using stacking ensemble,'' in \emph{2020 IEEE International Students' Conference on Electrical, Electronics and Computer Science}.\hskip 1em plus 0.5em minus 0.4em\relax IEEE, 2020, pp. 1--5.

\bibitem{JiewrongAcc2021}
J.~Chen, Y.~Jiang, Z.~Huang, X.~Guo, B.~Wu, L.~Sun, and T.~Wu, ``Fine-grained detection of driver distraction based on neural architecture search,'' \emph{IEEE Transactions on Intelligent Transportation Systems}, vol.~22, no.~9, pp. 5783--5801, 2021.

\bibitem{movieCLIP}
D.~Bose, R.~Hebbar, K.~Somandepalli, H.~Zhang, Y.~Cui, K.~Cole-McLaughlin, H.~Wang, and S.~Narayanan, ``Movieclip: Visual scene recognition in movies,'' in \emph{2023 IEEE/CVF Winter Conference on Applications of Computer Vision (WACV)}, 2023, pp. 2082--2091.

\bibitem{miech2019howto100m}
A.~Miech, D.~Zhukov, J.-B. Alayrac, M.~Tapaswi, I.~Laptev, and J.~Sivic, ``Howto100m: Learning a text-video embedding by watching hundred million narrated video clips,'' in \emph{Proceedings of the IEEE/CVF International Conference on Computer Vision}, 2019, pp. 2630--2640.

\bibitem{mithun2018webly}
N.~C. Mithun, R.~Panda, E.~E. Papalexakis, and A.~K. Roy-Chowdhury, ``Webly supervised joint embedding for cross-modal image-text retrieval,'' in \emph{Proceedings of the 26th ACM International Conference on Multimedia}, 2018, pp. 1856--1864.

\bibitem{dong2019dual}
J.~Dong, X.~Li, C.~Xu, S.~Ji, Y.~He, G.~Yang, and X.~Wang, ``Dual encoding for zero-example video retrieval,'' in \emph{Proceedings of the IEEE/CVF Conference on Computer Vision and Pattern Recognition}, 2019, pp. 9346--9355.

\bibitem{3dcnn}
S.~Xie, C.~Sun, J.~Huang, Z.~Tu, and K.~Murphy, ``Rethinking spatiotemporal feature learning: Speed-accuracy trade-offs in video classification,'' in \emph{Proceedings of the European Conference on Computer Vision (ECCV)}, September 2018.

\bibitem{syndd1}
M.~S. Rahman, A.~Venkatachalapathy, A.~Sharma, J.~Wang, S.~V. Gursoy, D.~Anastasiu, and S.~Wang, ``Synthetic distracted driving (syndd1) dataset for analyzing distracted behaviors and various gaze zones of a driver,'' \emph{Data in Brief}, vol.~46, p. 108793, 2023.

\bibitem{statefarm}
\BIBentryALTinterwordspacing
{Kaggle}, ``State farm distracted driver detection.'' [Online]. Available: \url{https://www.kaggle.com/competitions/state-farm-distracted-driver-detection/data}
\BIBentrySTDinterwordspacing

\bibitem{kingma2014adam}
D.~P. Kingma and J.~Ba, ``Adam: A method for stochastic optimization,'' \emph{arXiv preprint arXiv:1412.6980}, 2014.

\bibitem{aicity23top}
W.~Zhou, Y.~Qian, Z.~Jie, and L.~Ma, ``Multi view action recognition for distracted driver behavior localization,'' in \emph{2023 IEEE/CVF Conference on Computer Vision and Pattern Recognition Workshops (CVPRW)}, 2023, pp. 5375--5380.

\bibitem{un-aware}
A.~Roitberg, K.~Peng, D.~Schneider, K.~Yang, M.~Koulakis, M.~Martinez, and R.~Stiefelhagen, ``Is my driver observation model overconfident? input-guided calibration networks for reliable and interpretable confidence estimates,'' \emph{IEEE Transactions on Intelligent Transportation Systems}, vol.~23, no.~12, pp. 25\,271--25\,286, 2022.

\end{thebibliography}
 

\vspace{11pt}
\begin{IEEEbiography}[{\includegraphics[width=1in,height=1.25in,clip,keepaspectratio]{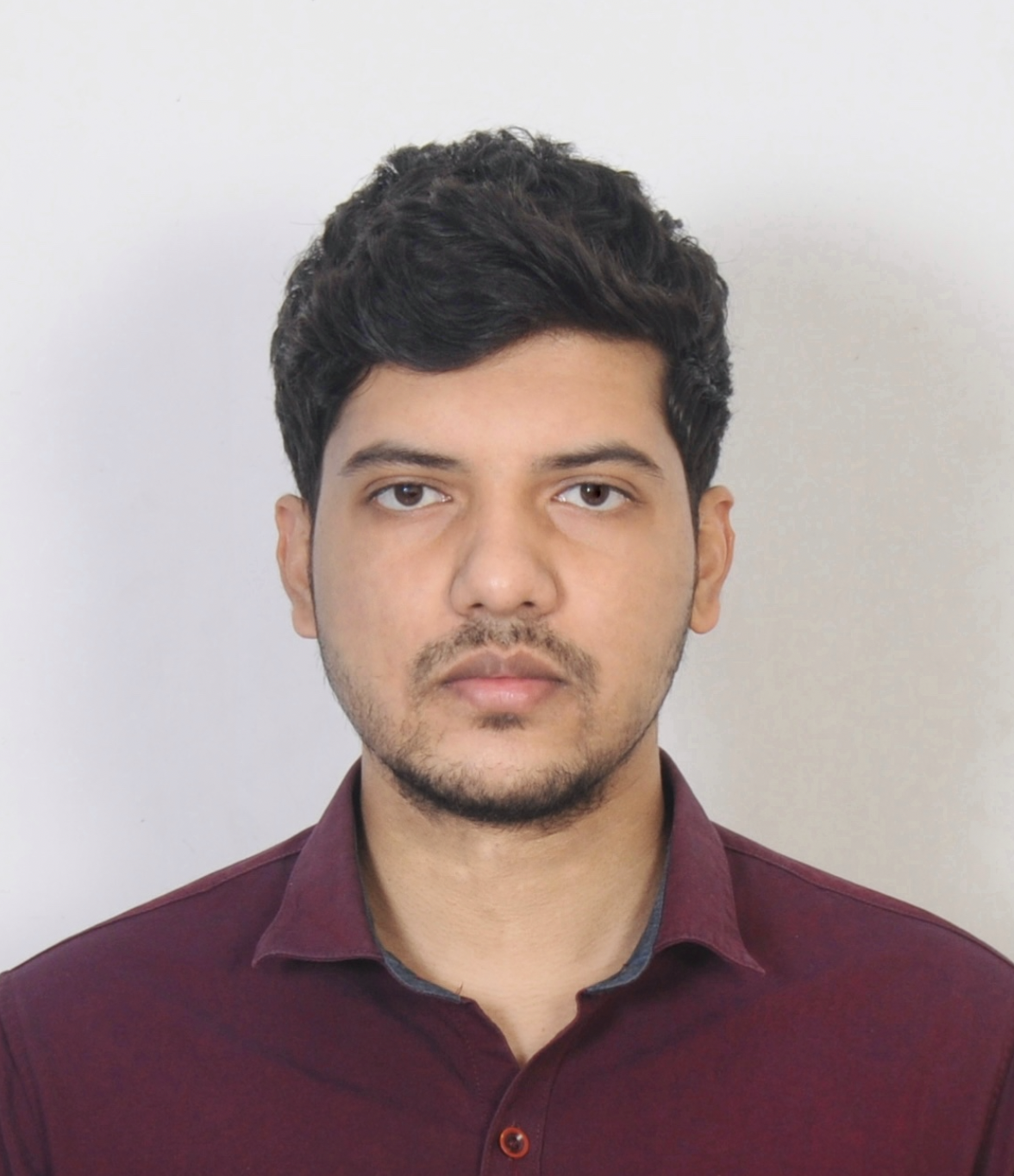}}]{Md Zahid Hasan}
received his B.S. degree in Electrical and Electronic Engineering from Bangladesh University of Engineering and Technology in 2018. He is currently pursuing his Ph.D. in Electrical Engineering with the Department of Electrical and Computer Engineering at Iowa State University, IA. His research interests include Vision-Language models, computer vision, machine learning and decentralized deep learning.
\vspace{-1cm}
\end{IEEEbiography}

\begin{IEEEbiography}[{\includegraphics[width=1in,height=1.25in,clip,keepaspectratio]{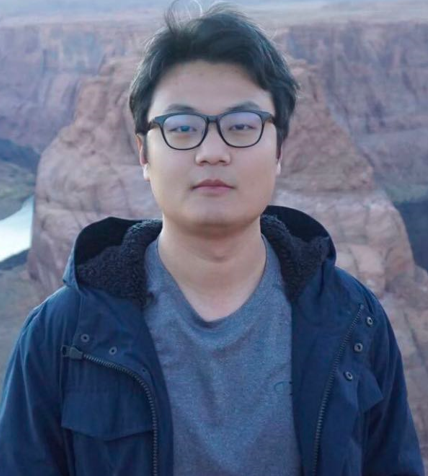}}]{Jiajing Chen}
received the B.S. degree in mechanical engineering from WuHan Institute of Technology, Wuhan, China in 2017 and M.S. degree in mechanical engineering from Syracuse University, Syracuse, NY, USA in 2019. He is currently pursuing the Ph.D. degree with the Department of Electrical Engineering and Computer Science, Syracuse University. His research interests include point cloud segmentation, weakly supervised object detection and few-shot learning.
\vspace{-1cm}
\end{IEEEbiography}

\begin{IEEEbiography}[{\includegraphics[width=1in,height=1.25in,clip,keepaspectratio]{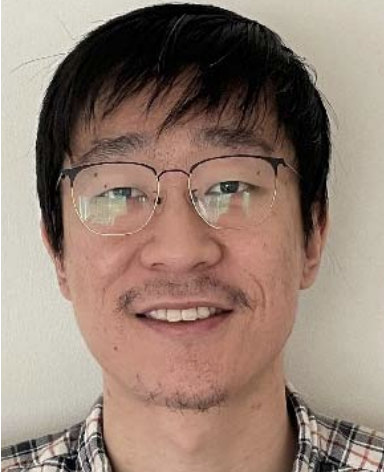}}]{Jiyang Wang}
received his Bachelor’s degree in electrical
engineering from Anhui University of Science
and Technology in 2017 and a Master’s degree in
electrical engineering from Syracuse University in
2019. He is currently a Ph.D. student in Electrical
and Computer Engineering at Syracuse University.
His research interests include computer vision, deep
learning and human computer interaction.
\vspace{-1cm}
\end{IEEEbiography}

\begin{IEEEbiography}[{\includegraphics[width=1in,height=1.25in,clip,keepaspectratio]{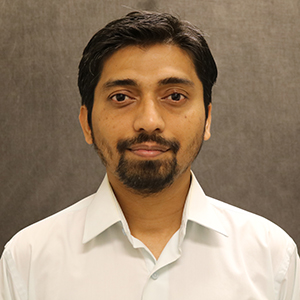}}]{Mohammed Shaiqur Rahman}
received his B.E. in Computer Science and Engineering from the National Institute of Engineering, India, in 2011. Then he joined IBM India Pvt Ltd and worked as a system engineer until early 2016. Later he joined the Department of Computer Science at Iowa State University, IA, where he completed his master’s degree, and he is currently pursuing his Ph.D. His research interests include Artificial Intelligence, Machine Learning, Computer Vision, and Federated Learning.
\vspace{-1cm}
\end{IEEEbiography}

\begin{IEEEbiography}[{\includegraphics[width=1in,height=1.25in,clip,keepaspectratio]{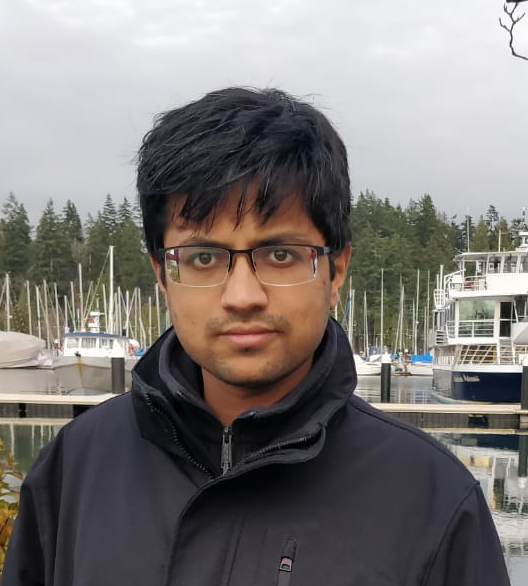}}]{Ameya Joshi}
received his B.S. degree in Electrical and Electronics Engineering from BITS Pilani, Goa, India in 2014. He is currently pursuing his Ph.D. in Electrical Engineering with the Department of Electrical and Computer Engineering at New York University. He was previously a Ph.D. student at Iowa State University (2018-2019) before going to NYU. His research interests include robust algorithms for multimodal models, adversarial robustness for deep learning, Vision-Text models, generative models for structured data, Physics Informed Generative models, Generative Adversarial Networks and computer vision.
\vspace{-1cm}
\end{IEEEbiography}

\begin{IEEEbiography}[{\includegraphics[width=1in,height=1.25in,clip,keepaspectratio]{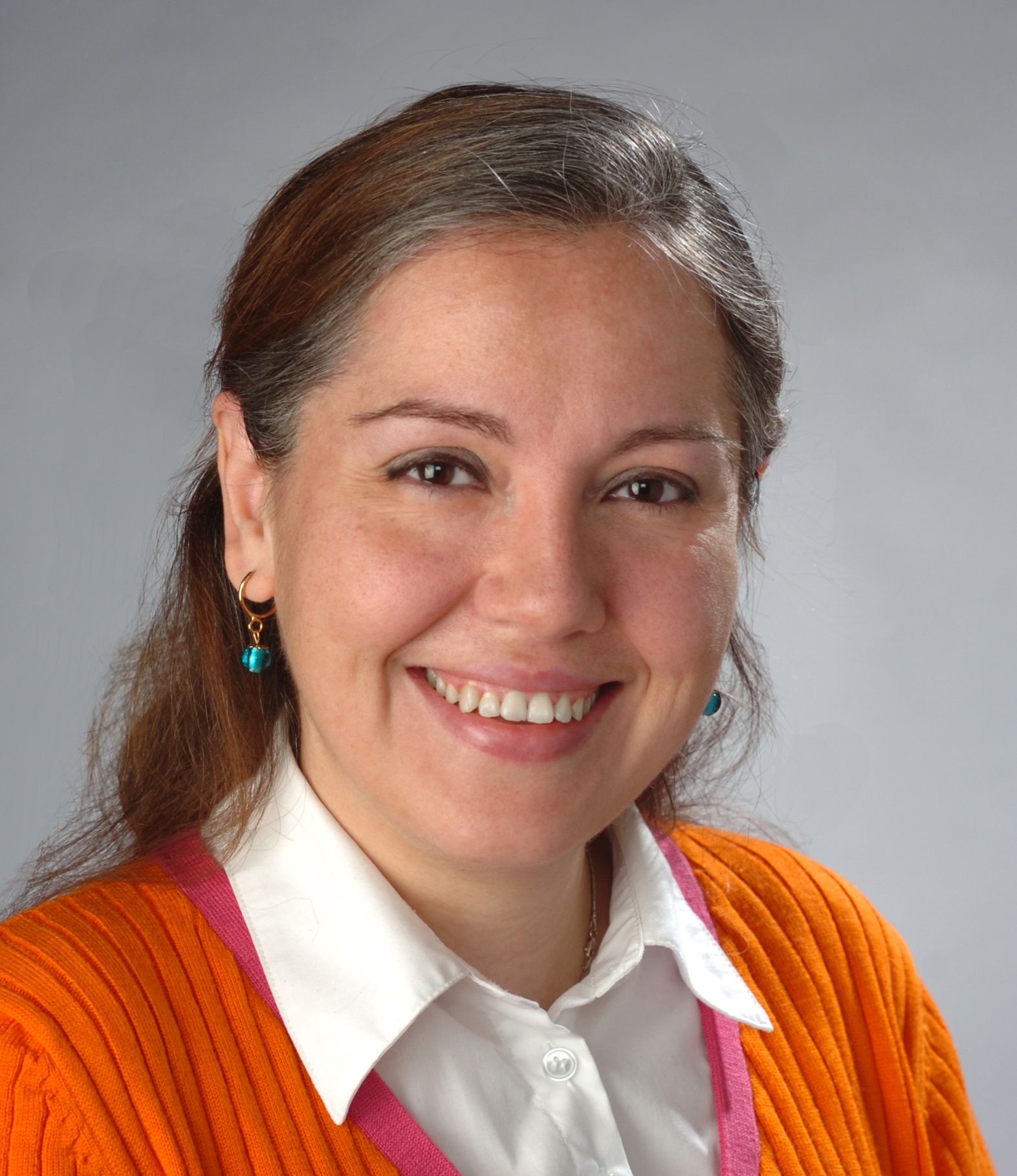}}]{Senem Velipasalar}(M'04--SM'14) received the Ph.D. and M.A degrees in electrical engineering from Princeton University, Princeton, NJ, USA, in 2007 and 2004, respectively, the M.S. degree in electrical sciences and computer engineering from Brown University, Providence, RI, USA, in 2001, and the B.S. degree in electrical and electronic engineering from Bogazici University, Istanbul, Turkey, in 1999. From 2007 to 2011, she was an Assistant Professor at the Department of Electrical Engineering, University of Nebraska-Lincoln. She is currently a Professor in the Department of Electrical Engineering and Computer Science, Syracuse University.
The focus of her research has been on machine learning, mobile camera applications, wireless embedded smart cameras, multicamera tracking and surveillance systems.
\vspace{-1cm}
\end{IEEEbiography}

\begin{IEEEbiography}[{\includegraphics[width=1in,height=1.25in,clip,keepaspectratio]{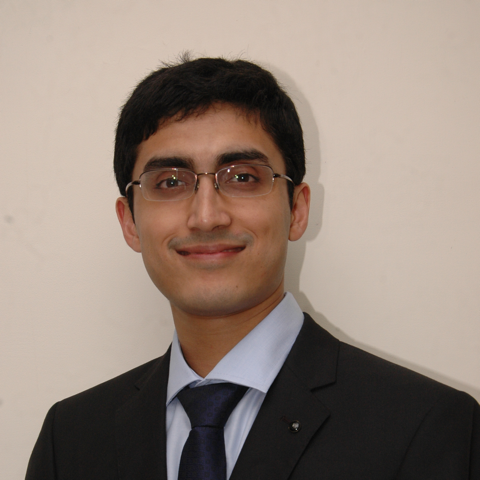}}]{Chinmay Hegde}
received his Ph.D. and M.S. in Electrical and Computer Engineering from Rice University in 2012 and 2010 respectively. He joined New York University, Tandon School of Engineering as an Assistant Professor in 2019. Previously, he was an Assistant Professor in the Department of Electrical and Computer Engineering at Iowa State University from 2015 to 2019. Prior to joining Iowa State, he was a postdoctoral associate in the Theory of Computation group at MIT from 2012 to 2015. His research focuses on developing principled, fast, and robust algorithms for diverse problems in machine learning, with applications to imaging and computer vision, materials design and transportation.

\vspace{-1cm}
\end{IEEEbiography}

\begin{IEEEbiography}[{\includegraphics[width=1in,height=1.25in,clip,keepaspectratio]{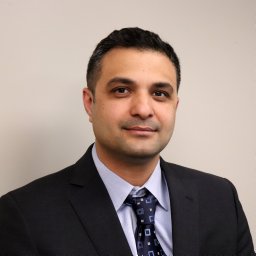}}]{Anuj Sharma}
received his Ph.D. degree in Civil Engineering from Purdue University in 2008, and his M.S. degree in Civil Engineering from Texas A\&M University in 2004. He was an Assistant Professor in the Department of Civil Engineering at the University of Nebraska-Lincoln. He is currently a Pitts-Des Moines Inc. Professor in Civil Engineering at Iowa State University. His research focuses on traffic operations, big data analytics, machine learning, and traffic safety.
\vspace{-1cm}
\end{IEEEbiography}

\begin{IEEEbiography}[{\includegraphics[width=1in,height=1.25in,clip,keepaspectratio]{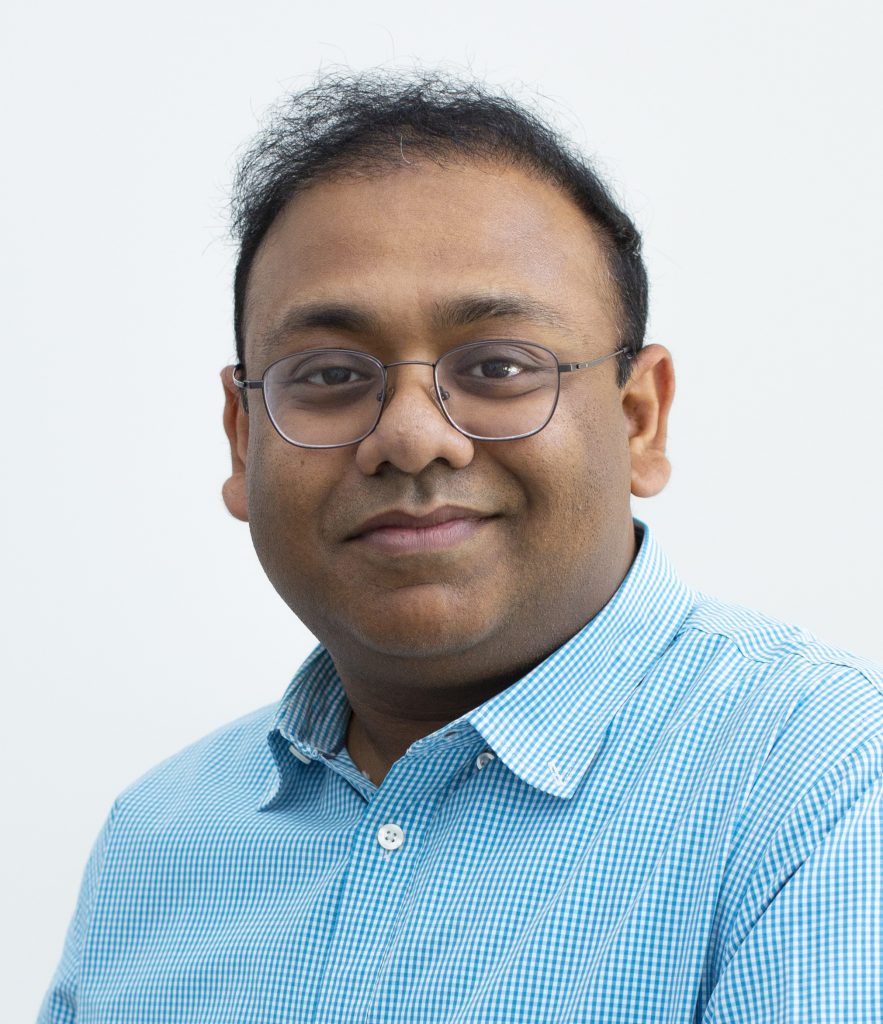}}]{Soumik Sarkar} received his Ph.D. in Mechanical Engineering from Penn State University in 2011. He joined the Department of Mechanical Engineering at Iowa State as an Assistant Professor in Fall 2014. Previously, he was with the Decision Support and Machine Intelligence group at the United Technologies Research Center for 3 years as a Senior Scientist. He is currently a Professor of Mechanical Engineering and Walter W Wilson Faculty Fellow in Engineering at Iowa State University. He also serves as the Director of Translational AI Center at Iowa State. His research focuses on machine learning, and controls for Cyber-Physical Systems applications such as aerospace, energy, transportation, manufacturing and agricultural systems. 
\vspace{-1cm}
\end{IEEEbiography}
\vfill

\end{document}